\documentclass[letterpaper,journal]{IEEEtran}
\usepackage[pdftex]{graphicx}
\usepackage{amsmath,amssymb,amsopn,amstext,amsfonts}
\usepackage{xcolor}
\usepackage{ulem}
\usepackage{lipsum}
\usepackage{times}
\usepackage{cancel}
\usepackage[space]{cite}
\usepackage{pdfsync}
\usepackage{color}
\usepackage{algorithm}
\usepackage{algorithmicx}
\usepackage{algpseudocode}
\usepackage{bm}
\usepackage{diagbox}
\usepackage{url}
\usepackage{float}
\usepackage{amsthm}

\usepackage{verbatim}
\usepackage{graphicx}
\usepackage{makecell}
\usepackage{array,xcolor,colortbl}
\usepackage{multirow}
\usepackage{mathtools}
\usepackage[linkcolor=black,citecolor=black,urlcolor=black,colorlinks=true]{hyperref}
\usepackage{orcidlink}
\usepackage{colortbl}
\makeatletter
\def\tagform@#1{\maketag@@@{\normalsize(#1)\@@italiccorr}}
\makeatother
\usepackage{amsthm}

\usepackage{multirow}
\usepackage{booktabs}
\graphicspath{{./}}
\DeclareGraphicsExtensions{.pdf,.png,.jpg,.eps,.svg}
\IEEEoverridecommandlockouts
\newcommand{\Para}[1]{\par}
\title{Aerodynamic Prior-Free Coordinated Trajectory Generation and Tracking Control for a Tail-Sitter UAV}
\author{\mbox{Erchao Rong\,\orcidlink{0009-0008-4701-9127}}, \mbox{Zihao Liu\,\orcidlink{0009-0001-5523-3229}}, \mbox{Junning Liang\,\orcidlink{0009-0008-2608-9310}}, \mbox{Jianguo Wang\,\orcidlink{0009-0004-1208-1300}}, \mbox{Xiao Jie\,\orcidlink{0009-0004-6562-9800}}, \mbox{Haoran Fu\,\orcidlink{0009-0004-0003-032X}}, \mbox{Ziliang Chen\,\orcidlink{0000-0003-2026-6028}}, \mbox{Ximin Lyu\,\orcidlink{0000-0002-5204-5628}}
    \thanks{This work was supported in part by the National Key Research and Development Program of China under Grant 2023YFB4706600, in part by the National Natural Science Foundation of China under Grant 62303495, in part by the Major Key Project of PCL (PCL2025A02), and in part by the Young Talent Support Project of Guangzhou Association for Science and Technology under Grant QT-2025-004. (Erchao Rong and Zihao Liu contributed equally to this work.) (Corresponding author: Ximin Lyu.)}
    \thanks{Erchao Rong, Zihao Liu, Junning Liang, Jianguo Wang, Xiao Jie, and Haoran Fu are with the School of Intelligent Systems Engineering, Sun Yat-sen University, Guangzhou 510275, China (e-mail: \href{mailto:rongerch@outlook.com}{rongerch@outlook.com}; \href{mailto:liuzh297@gmail.com}{liuzh297@gmail.com}; \href{mailto:gordonliang27@foxmail.com}{gordonliang27@foxmail.com}; \href{mailto:jgw_2022@163.com}{jgw\_2022@163.com}; \href{mailto:jiexiao_2025@163.com}{jiexiao\_2025@163.com}; \href{mailto:fu.haoran@outlook.com}{fu.haoran@outlook.com}).}
    \thanks{Ziliang Chen is with Pengcheng Laboratory, Shenzhen 518055, China (e-mail: \href{mailto:c.ziliang@yahoo.com}{c.ziliang@yahoo.com}).}
    \thanks{Ximin Lyu is with the School of Intelligent Systems Engineering, Sun Yat-sen University, Guangzhou 510275, China, Pengcheng Laboratory, Shenzhen 518055, China, and the Differential Robotics Technology Company, Ltd., Hangzhou 311100, China (e-mail: \href{mailto:lvxm6@mail.sysu.edu.cn}{lvxm6@mail.sysu.edu.cn}).}
}

\newcommand{\xee}[1]{\lfloor#1\rfloor}
\newcommand{\norm}[1]{\left\lVert#1\right\rVert}

\newcommand{\stateYb}{\bm{y}_b}
\newcommand{\stateXb}{\bm{x}_b}
\newcommand{\stateZb}{\bm{z}_b}

\newcommand{\stateFa}{\bm{f}_a}

\newcommand{\stateFT}{\bm{f}_T}

\newcommand{\stateFTBx}{\prescript{\mathcal{B}}{}{\bm{f}_{T_x}}}
\newcommand{\stateAp}{\bm{a}_p}
\newcommand{\stateApS}{\prescript{\mathcal{S}}{}{\bm{a}_p}}

\newcommand{\stateApBx}{\prescript{\mathcal{B}}{}{\bm{a}_{p_x}}}

\newcommand{\stateApBz}{\prescript{\mathcal{B}}{}{\bm{a}_{p_z}}}
\newcommand{\stateApSx}{\prescript{\mathcal{S}}{}{\bm{a}_{p_x}}}
\newcommand{\stateApSy}{\prescript{\mathcal{S}}{}{\bm{a}_{p_y}}}
\newcommand{\stateApSz}{\prescript{\mathcal{S}}{}{\bm{a}_{p_z}}}
\newcommand{\stateApDotSx}{\prescript{\mathcal{S}}{}{\dot{\bm{a}}_{p_x}}}
\newcommand{\stateApDotSy}{\prescript{\mathcal{S}}{}{\dot{\bm{a}}_{p_y}}}
\newcommand{\stateApDotSz}{\prescript{\mathcal{S}}{}{\dot{\bm{a}}_{p_z}}}
\newcommand{\stateApDotS}{\prescript{\mathcal{S}}{}{\dot{\bm{a}}_{p}}}

\newcommand{\stateThrustForceBx}{\prescript{\mathcal{B}}{}{\bm{f}_{T_x}}}

\newcommand{\stateOmegaB}{\prescript{\mathcal{B}}{}{\bm{\Omega}_b}}

\newcommand{\stateDotOmegaB}{\prescript{\mathcal{B}}{}{\dot{\bm{\Omega}}_b}}

\newcommand{\stateOmegaS}{\prescript{\mathcal{S}}{}{\bm{\Omega}_s}}
\newcommand{\stateOmegaSx}{\prescript{\mathcal{S}}{}{\bm{\Omega}_{s_x}}}
\newcommand{\stateOmegaSy}{\prescript{\mathcal{S}}{}{\bm{\Omega}_{s_y}}}
\newcommand{\stateOmegaSz}{\prescript{\mathcal{S}}{}{\bm{\Omega}_{s_z}}}
\newcommand{\costMPC}{\mathcal{J}_{\text{MPC}}}
\newcommand{\aeroMoment}{\bm{M}_a}
\newcommand{\DotCz}{\prescript{\mathcal{B}}{}{\dot{\bm{c}}_z}}
\newcommand{\aBx}{\prescript{\mathcal{B}}{}{\bm{a}_x}}
\newcommand{\aBy}{\prescript{\mathcal{B}}{}{\bm{a}_y}}
\newcommand{\aBz}{\prescript{\mathcal{B}}{}{\bm{a}_z}}
\newcommand{\stateOmegaBx}{\prescript{\mathcal{B}}{}{\bm{\Omega}_{b_x}}}
\newcommand{\stateOmegaBy}{\prescript{\mathcal{B}}{}{\bm{\Omega}_{b_y}}}
\newcommand{\stateOmegaBz}{\prescript{\mathcal{B}}{}{\bm{\Omega}_{b_z}}}

\newcommand{\stateMomentTBx}{\prescript{\mathcal{B}}{}{\boldsymbol{M}_{T_x}}}
\newcommand{\stateMomentTBy}{\prescript{\mathcal{B}}{}{\boldsymbol{M}_{T_y}}}
\newcommand{\stateMomentTBz}{\prescript{\mathcal{B}}{}{\boldsymbol{M}_{T_z}}}
\newcommand{\stateMomentT}{\prescript{\mathcal{B}}{}{\boldsymbol{M}_T}}
\newcommand{\fc}{\bm{f}_c}
\newcommand{\stateMomentB}{\prescript{\mathcal{B}}{}{\boldsymbol{M}}}

\newcommand{\stateMomentAero}{
\prescript{\mathcal{B}}{}{\boldsymbol{M}_a}
}

\newcommand{\stateAeroCoefz}{\prescript{\mathcal{B}}{}{\bm{c}_z}}
\newcommand{\stateAeroForce}{\bm{f}_{a}}
\newcommand{\stateAeroForceB}{\prescript{\mathcal{B}}{}{\bm{f}_{a}}}
\newcommand{\stateAeroForceBx}{\prescript{\mathcal{B}}{}{\bm{f}_{a_x}}}
\newcommand{\stateAeroForceBy}{\prescript{\mathcal{B}}{}{\bm{f}_{a_y}}}
\newcommand{\stateAeroForceBz}{\prescript{\mathcal{B}}{}{\bm{f}_{a_z}}}
\newcommand{\cx}{\prescript{\mathcal{B}}{}{\bm{c}_x}}
\newcommand{\cy}{\prescript{\mathcal{B}}{}{\bm{c}_y}}
\newcommand{\cz}{\prescript{\mathcal{B}}{}{\bm{c}_z}}

\newcommand{\stateAeroForceSy}{\prescript{\mathcal{S}}{}{\bm{f}_{a_y}}}

\newcommand{\SOthree}{\text{SO(3)}}

\newcommand{\stateP}{\bm{p}}
\newcommand{\stateV}{\bm{v}}
\newcommand{\stateVtranspose}{\bm{v}^{\text{T}}}
\newcommand{\stateA}{\bm{a}}
\newcommand{\stateJ}{\bm{j}}
\newcommand{\stateG}{\bm{g}}
\newcommand{\stateR}{\mathbf{R}_{\mathcal{B}}}
\newcommand{\stateDotR}{\dot{\mathbf{R}}_{\mathcal{B}}}

\newcommand{\xb}{\bm{x}_b}
\newcommand{\yb}{\bm{y}_b}
\newcommand{\zb}{\bm{z}_b}
\newcommand{\xs}{\bm{x}_s}
\newcommand{\xstranpose}{\bm{x}_s^{\text{T}}}
\newcommand{\ys}{\bm{y}_s}
\newcommand{\zs}{\bm{z}_s}
\newcommand{\xw}{\bm{x}_w}
\newcommand{\yw}{\bm{y}_w}
\newcommand{\zw}{\bm{z}_w}
\newcommand{\Bvx}{\prescript{\mathcal{B}}{}{\bm{v}_x}}

\newcommand{\Bvy}{\prescript{\mathcal{B}}{}{\bm{v}_y}}
\newcommand{\Bvz}{\prescript{\mathcal{B}}{}{\bm{v}_z}}

\newcommand{\RS}{\mathbf{R}_{\mathcal{S}}}
\newcommand{\RB}{\mathbf{R}_{\mathcal{B}}}
\newcommand{\SRB}{\prescript{\mathcal{S}}{}{\mathbf{R}_{\mathcal{B}}}}
\newcommand{\BRS}{\prescript{\mathcal{B}}{}{\mathbf{R}_{\mathcal{S}}}}
\newcommand{\dotAlpha}{\dot{\alpha}}
\newcommand{\Bfa}{\prescript{\mathcal{B}}{}{\bm{f}_a}}
\newcommand{\Bfax}{\prescript{\mathcal{B}}{}{\bm{f}_{a_x}}}
\newcommand{\Bfay}{\prescript{\mathcal{B}}{}{\bm{f}_{a_y}}}
\newcommand{\Bfaz}{\prescript{\mathcal{B}}{}{\bm{f}_{a_z}}}

\newcommand{\fa}{\bm{f}_a}

\newcommand{\refyb}{\bm{y}_b^{\text{ref}}}
\newcommand{\refpos}{\bm{p}^{\text{ref}}}
\newcommand{\refvel}{\bm{v}^{\text{ref}}}
\newcommand{\refacc}{\bm{a}_p^{\text{ref}}}

\newcommand{\ctrlat}{{a_T^{\text{c}}}}
\newcommand{\ctrlOmegaB}{\prescript{\mathcal{B}}{}{\bm{\Omega}_b^{\text{c}}}}
\newcommand{\ctrlApx}{{\prescript{\mathcal{B}}{}{\bm{a}_{p_x}^{\text{c}}}}}

\newcommand{\RR}{\mathbb{R}}

\newcommand{\costPlanning}{\mathcal{J}_{\text{PLN}}}
\newcommand{\costOriginal}{\mathcal{J}_{\text{gcopter}}}
\newcommand{\costConstr}{\mathcal{J}_{\text{feas}}}
\newcommand{\CDflat}{C_{D,\text{flat}}}

\newcommand{\dfForward}{\mathfrak{D}_\mathcal{F}}

\newcommand{\plnMaxOmega}{\prescript{\mathcal{B}}{}{\bm{\Omega}_{b,\text{max}}}}

\newcommand{\plnMaxTanAcc}{\prescript{S}{}{\bm{a}}_{x,\text{max}}}
\newcommand{\TanAcc}{\prescript{S}{}{\bm{a}}_{x}}
\newcommand{\plnMinTanAcc}{\prescript{S}{}{\bm{a}}_{x,\text{min}}}
\newcommand{\plnMaxVel}{V_{\text{max}}}
\newcommand{\dfBackward}{\mathfrak{D}_{\mathcal{B}}}
\newcommand{\MINCO}{\mathfrak{T}_{\text{MINCO}}}

\newcommand{\refposTraj}{\bm{p}^{\text{ref}}(t)}
\newcommand{\refvelTraj}{\bm{v}^{\text{ref}}(t)}
\newcommand{\refaccTraj}{\bm{a}^{\text{ref}}(t)}
\newcommand{\refybTraj}{\bm{y}_b^{\text{ref}}(t)}
\newcommand{\trajDecWaypoints}{\bm{Q}}
\newcommand{\trajDecTimeAllocation}{\bm{T}}
\newcommand{\Givens}[1]{\textbf{Given:}~#1}

\begin{document}
\maketitle
\begin{abstract}
This paper presents a coordinated trajectory generation and tracking control framework for a tail-sitter unmanned aerial vehicle (UAV), which does not require aerodynamic priors identified for a specific airframe while addressing the challenge of flight control under highly nonlinear aerodynamics across the full flight envelope.
The core innovation lies in employing phase-specific aerodynamic modeling strategies for planning and tracking, tailored to their distinct functional characteristics, without requiring airframe-specific aerodynamic priors.
Specifically, the $\phi$-theory model under coordinated flight is employed to derive an analytic differential flatness mapping, and a simplified but locally accurate model is established for predictive control to enable real-time aerodynamic parameter estimation.
The proposed framework is evaluated extensively through both simulation and challenging real-world flight tests under mild wind conditions, showing high-precision tracking and adaptability across the tested aerodynamic conditions.
To the best of our knowledge, this is the first real-world demonstration of accurate trajectory tracking over tested flight regimes spanning the full envelope of a tail-sitter UAV without relying on aerodynamic identification campaigns. The source code of our framework is available at: \href{https://github.com/SYSU-HILAB/AP-PnC}{https://github.com/SYSU-HILAB/AP-PnC}.
\end{abstract}
\section{Introduction}
\Para{intro-01}
\IEEEPARstart {T}ail-sitter UAVs are characterized by their mechanical simplicity and lightweight structure, compared to other hybrid vertical take-off and landing (VTOL) UAVs.
However, the large attitude shifts required for transition cause significant variations in the angle of attack (AoA).
Such complex and nonlinear aerodynamics render trajectory planning and tracking control exceptionally challenging, thereby limiting the widespread application of tail-sitter UAVs.

\Para{intro-02}
Current high-performance studies~\cite{lu2024trajectory, lu2025autonomous} in tail-sitter control have been largely built upon high-fidelity aerodynamic models. While these approaches have demonstrated impressive performance, their practical deployment often faces several hurdles: 1) controllers are optimized for a specific airframe, hindering portability; 2) accurate aerodynamic estimation typically entails sophisticated software and hardware setups; 3) identification of a reliable aerodynamic model is a resource-intensive process.
To streamline deployment, this work introduces an integrated flight framework that obviates costly aerodynamic model identification, enabling broader tail-sitter adoption.
\Para{intro-03}
Standard tail-sitter tracking controllers typically employ a cascaded architecture with an outer position loop and an inner attitude loop.
The primary challenge lies in reactively mapping the desired acceleration to an attitude command.
Early methods~\cite{eth-full, lyu-cas} empirically tuned predefined angular velocity profiles to balance flight feasibility and performance, necessitating extensive manual effort.
To automate this process, later optimization approaches~\cite{zhou-scp-att, eth-flywing, quanquan-liftwing} explicitly model aerodynamic forces to compute feasible attitude commands.
Yet, a trade-off exists: high-fidelity models are computationally prohibitive~\cite{zhou-scp-att}, while efficient low-fidelity models often lack sufficient accuracy~\cite{eth-flywing, quanquan-liftwing}.
To free the control design from this efficacy-portability dilemma, sensor-based paradigms have been explored.
Incremental nonlinear dynamic inversion (INDI) techniques~\cite{indi-smeur, mit_estimation, rohr2024unified} and a linear acceleration model~\cite{xuwei-acc, RAL-ducted-fan} were introduced to compensate for modeling errors using real-time sensor estimates, offering a more lightweight identification burden regarding aerodynamics and actuators.
Such instantaneously estimated aerodynamic forces from accelerometers are only utilized reactively, without predictive exploitation.
\Para{intro-04}
To transition from reactive compensation to proactive adaptation, model predictive control (MPC) employs online explicit optimization over receding short horizons, whereas learning-based methods implicitly encode foresight priors into a reactive policy via offline training~\cite{song2023reaching}.
For racing drones, which also suffer from complex aerodynamic effects, recent works utilizing model-free learning~\cite{kaufmann2023champion}, model-based learning~\cite{skydreamer2025}, and differentiable physics~\cite{differentiable-physics} have demonstrated impressive performance. While their success hinges on high-fidelity simulators and elaborate learning pipelines to bridge the sim-to-real gap, such infrastructures remain largely unavailable for tail-sitter platforms.
Consequently, the tail-sitter community continues to rely predominantly on MPC-based solutions grounded in explicitly formulated aerodynamic models.
Yet, the real-time onboard optimization precludes the direct use of a full nonlinear aerodynamic model. To address this issue, Lu \textit{et al.}~\cite{lu2024trajectory, lu2025autonomous} linearize the system dynamics along a reference trajectory to derive a time-varying error-state model.
Crucially, their linearization process necessitates a high-fidelity aerodynamic model to establish the reference trajectory via differential flatness. To date, achieving predictive control performance without requiring advance aerodynamic characterization remains an open challenge.
\Para{intro-05}
Beyond tracking control, trajectory planning remains heavily dependent on high-fidelity aerodynamic models.
Trajectory planning without differential flatness for winged aircraft often relies on optimization within the joint state-control space.
Common approaches include direct transcription methods~\cite{pla-dic-collocation}, discretization via motion primitives~\cite{mp_library_2019,mp_library_2020}, or the use of population-based metaheuristics~\cite{turkey2023, turkey2023uav}.
However, these methods struggle to balance optimality and computational tractability as the problem scales up.
Alternatively, Wang \textit{et al.}~\cite{minco2022} proposed a trajectory planning framework in the flat output space, which achieves a linear computational complexity relative to the number of trajectory segments.
Building on this, Yu \textit{et al.}~\cite{yu2025top} leveraged a closed-form differential flatness to enable the parallel generation of all segments, thereby achieving constant-time complexity.
While winged aircraft in coordinated flight are proven to be differentially flat~\cite{df_co}, mapping these flat outputs to body-frame angular velocity remains a bottleneck. This process requires solving for the AoA through aerodynamic models, which generally leads to two compromises: 1) \textit{Numerical dependency}: Using a high-fidelity model often necessitates an optimization subroutine to determine the AoA~\cite{lu2024trajectory}.
Rather than allowing for point-wise algebraic mapping, this approach requires strictly serial optimization. The dependency of each time step on its predecessor prevents the use of parallel computing.
2) \textit{Model simplification}: Existing algebraic solutions often rely on restrictive assumptions that limit their applicability. The derivation in ~\cite{df_co} is confined to stability-frame angular velocity. Other methods employ low-alpha approximations~\cite{df-low}, 2-D planar dynamics~\cite{df-longitudinal}, or a reduced $\phi$-theory model valid for a limited range of AoA~\cite{df-lift-wing}.
Furthermore, Tal \textit{et al.}~\cite{mit_traj2022} utilized a $\phi$-theory model assuming negligible lateral forces to generate uncoordinated maneuver trajectories.
While effective for their specific platform, this assumption does not hold for general tail-sitters, where lateral aerodynamics play a critical role during such maneuvers.
    Therefore, a versatile trajectory planning method free of such dependencies remains absent. This deficiency, combined with the restrictive reliance on models in tracking control, highlights the lack of a fully integrated flight framework that operates without airframe-specific aerodynamic priors.
\Para{intro-06}
To tackle this challenge, we present an integrated methodology with the following contributions:
\begin{enumerate}
    \item An analytic expression for the differential flatness mapping for tail-sitters under coordinated flight, based on the full $\phi$-theory model without simplification. This formulation can be leveraged to generate feasible trajectories and accurate reference states without requiring advance aerodynamic characterization.
    \item A novel nonlinear MPC with real-time aerodynamic parameter estimation. This controller continuously identifies and adapts to the aircraft's aerodynamic characteristics in flight, thereby obviating costly aerodynamic model identification.
    \item Open-source code to facilitate further validation and investigation by the community.
\end{enumerate}
To our knowledge, this is the first method to achieve sub-meter position RMSE over the tested flight regimes without relying on aerodynamic identification campaigns.
The framework assumes wind-free conditions, with failure modes characterized in Sec.~\ref{sec:op-bound}.
In the following sections, Sec.~\ref{sec:system-overiew} first outlines the flight dynamics and system framework. Subsequently, Sec.~\ref{sec:traj-planing} leverages the $\phi$-theory model to derive differential flatness for trajectory planning, which is then tracked by our proposed MPC in Sec.~\ref{sec:traj-tracking-controller}. We validate the effectiveness and robustness of the entire framework via experiments in Sec.~\ref{sec:exp} and conclude the work in Sec.~\ref{sec:conclusion}.
\section{System Overview}
\label{sec:system-overiew}
\begin{figure}[t]
    \centering
      \includegraphics[width=1.0\linewidth]{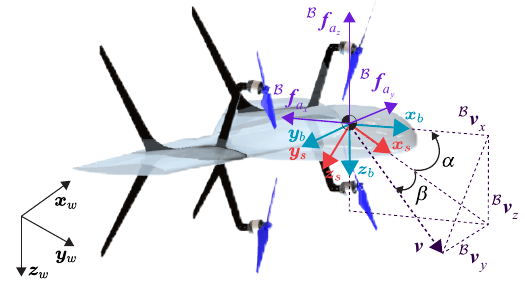}
    \caption{Coordinate frames: the inertial world frame $\mathcal{W}$, body frame $\mathcal{B}$ and stability frame $\mathcal{S}$}
    \label{fig:frames}
\end{figure}
\subsection{Coordinate Frames}
\Para{overview-01}
In this paper, we use $\mathbf{e}_x\!=\![1,0,0]^{\text{T}}$, $\mathbf{e}_y\!=\![0,1,0]^{\text{T}}$, and $\mathbf{e}_z\!=\![0,0,1]^{\text{T}}$ to represent the standard unit vectors along the $x$-, $y$-, and $z$-axes of a 3-D Cartesian coordinate system, respectively. An entity vector $\mathbf{r}$, represented in a specific coordinate frame $\mathcal{F}$, is denoted by $\prescript{{\mathcal{F}}}{}{\mathbf{r}}$.
The components of $\mathbf{r}$ in the frame are given by $
\prescript{\mathcal{F}}{}{\mathbf{r}_x}
\!=\!\mathbf{e}_x \cdot  \prescript{{\mathcal{F}}}{}{\mathbf{r}}$,
 $
\prescript{\mathcal{F}}{}{\mathbf{r}_y}\!
 =\!\mathbf{e}_y \cdot  \prescript{{\mathcal{F}}}{}{\mathbf{r}}$,
 and
 $
\prescript{\mathcal{F}}{}{\mathbf{r}_z}
 = \mathbf{e}_z \cdot  \prescript{{\mathcal{F}}}{}{\mathbf{r}}$
.
A rotation matrix $
 \prescript{{\mathcal{F}_0}}{}{ \mathbf{R}_{\mathcal{F}_1}
}$ represents the orientation of the frame $\mathcal{F}_1$ relative to the frame $\mathcal{F}_0$.
For brevity, the left superscript is omitted for quantities in the inertial frame.
\Para{overview-02}
The coordinate frames are shown in Fig.~\ref{fig:frames}. The inertial world frame $\mathcal{W}$ follows the north-east-down (NED) convention, with orthonormal basis $\{\xw, \yw, \zw\}$. The body-fixed frame $\mathcal{B}$ has its origin at the center of mass, with $\xb$ pointing forward and $\yb$ toward the right wing tip. The stability frame $\mathcal{S}$ is obtained by rotating $\mathcal{B}$ about $\yb$ by $-\alpha$, where the AoA $\alpha$ is defined as
\begin{equation}
    \label{eq:def-alpha}
    \alpha = \operatorname{atan2}\left( \Bvz,  \Bvx\right).
\end{equation}
The relation between the attitude of the stability frame $\RS$ and the attitude of the body frame $\RB$ is characterized as follows:
\begin{equation}
    \label{eq:transformtion-S-B}
    \stateR = \RS \SRB.
\end{equation}
As illustrated in Fig.~\ref{fig:frames}, the $\SRB$ is defined as
\begin{equation}
    \label{eq:express-SRB}
    \SRB =
        \begin{bmatrix}
        \cos \alpha & 0 & \sin \alpha \\
        0 & 1 & 0 \\
        -\sin \alpha & 0 & \cos \alpha
    \end{bmatrix}.
\end{equation}
Substituting \eqref{eq:express-SRB} into \eqref{eq:transformtion-S-B} will yield
\begin{equation}
    \label{eq:expansion-axes}
        [
            \xs \mid \ys \mid \zs
        ]
        =
        [
            \xb \cos \alpha + \zb \sin \alpha
            \mid
            \yb
            \mid
            -\xb \sin \alpha + \zb \cos \alpha
        ].
\end{equation}
\Para{overview-03}
The position of the tail-sitter is denoted as $\bm{p}$ and its derivatives, velocity, acceleration and jerk as $\stateV$, $\stateA$, $\stateJ$, respectively.
\subsection{Kinematics}
\Para{overview-04}
The translational kinematics are described by
\begin{equation}
    \label{eq:translational-kinematics}
    \begin{aligned}
        \dot{\stateP} & =\stateV, \\
        \dot{\stateV} & = \stateA = \stateG+ \stateAp, \\
    \end{aligned}
\end{equation}
where $\stateAp = \stateA - \stateG$ is the proper acceleration produced by the thrust force $\bm{f}_T$ and aerodynamic force $\bm{f}_a$, and it is given by
\begin{equation}
    \label{eq:ap-kinematics}
    \stateAp = \frac{\stateFa + \stateFT}{m}.
\end{equation}
The rotational kinematics are written as
\begin{subequations}
    \label{eq:rotational-kinematics}
    \begin{align}
        \label{eq:rotational-kinematics-R}
         \stateDotR &=   \stateR \xee{\stateOmegaB}, \\
        \label{eq:rotational-kinematics-Omega}
        \stateDotOmegaB&= \mathbf{J}^{-1}
        \left(
        \stateMomentB
        -
        \xee{\stateOmegaB}  \mathbf{J} \stateOmegaB
        \right),
    \end{align}
\end{subequations}
where $\mathbf{J}$ is the inertia matrix of the aircraft, $\stateOmegaB$ denotes the body-frame angular velocity. The notation $\xee{\cdot}$ denotes the map that transforms a vector to a skew-matrix, such that $\xee{\bm{a}} \bm{b} = \bm{a} \times \bm{b}, \forall \bm{a}, \bm{b} \in \mathbb{R}^3$. $\stateMomentB$ is the total moment acting on the vehicle,  produced by the aerodynamic moment $\stateMomentAero$ and the thrust moment $\stateMomentT$.
\Para{overview-05}
We define the nominal state of our system $\bm{x}$ as
\begin{equation}
    \label{eq:nominal-state}
    \bm{x} = \left(\stateP, \stateV, \stateR, \stateOmegaB \right) \in \RR^3 \times \RR^3 \times \SOthree \times \RR^3.
\end{equation}
The aerodynamic moment $\stateMomentAero$ is treated as a disturbance in this paper, and compensated by a low-level controller.
The dynamical modeling of $\bm{f}_a$, $\bm{f}_T$, and $\stateMomentT$
is handled differently during planning and tracking control.
\subsection{System Framework}
\begin{figure}[t]
    \centering
    \includegraphics[height=6cm, width=1.0\linewidth]{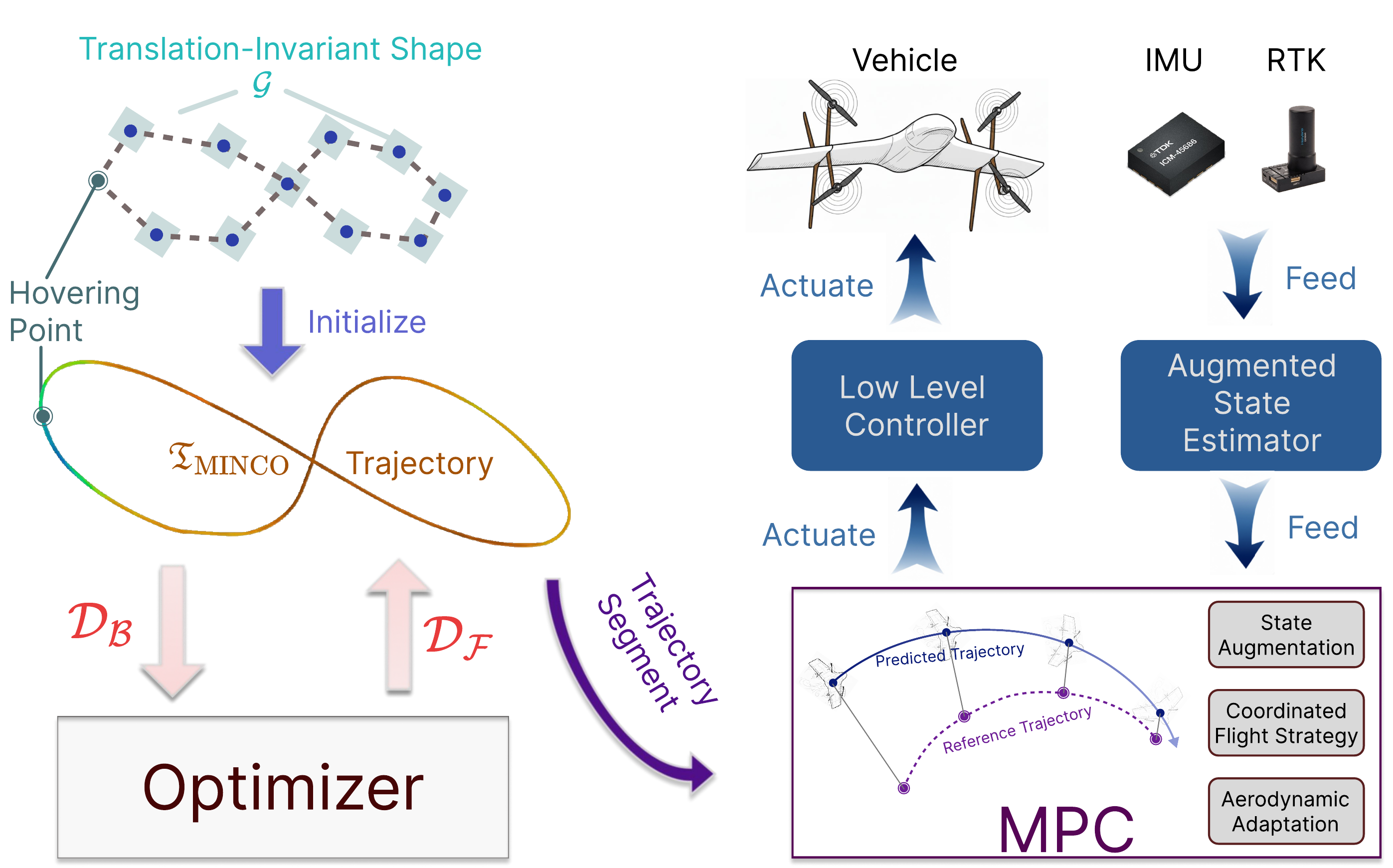}
    \caption{System framework.}
    \label{fig:framework}
\end{figure}

\Para{overview-06}
To accurately achieve an aerodynamic prior-free trajectory tracking, it is critical to recognize that the trajectory planning phase requires a simplified, feasibility-oriented model in long-term horizons, whereas the tracking phase demands a locally valid model in short-term horizons.
Specifically, the key lies in a subtle yet deliberate modeling of the tail-sitter's full-flight-envelope $\fa$ in the two phases.
Particularly in this paper, we extend the differential flatness property of the coordinated vehicle model~\cite{df_co} to dynamics-dependent terms---the body-frame angular velocity $\stateOmegaB$ and  the thrust acceleration $a_T$---leveraging the $\phi$-theory aerodynamic model~\cite{phi2019}.
Building on this flatness property and the flatness-based trajectory planning framework in~\cite{minco2022}, we formulate an aerodynamic prior-free trajectory planning optimization problem by parameterizing the $\phi$-theory model using a single flat-plate drag coefficient $\CDflat$, with a value of 1.28.
As for trajectory tracking, our MPC incorporates only a single nonlinear longitudinal aerodynamic parameter. This parameter is estimated online during flight and used to predict aerodynamic force variations over the prediction horizon.
\Para{overview-07}
The framework of our aerodynamic prior-free trajectory planning and tracking control is shown in Fig.~\ref{fig:framework}. As detailed in Sec.~\ref{sec:traj-planing}, a dynamically feasible trajectory is first generated offline by our trajectory planner. The reference position, velocity and lateral axis vectors are extracted based on the differential flatness property of the tail-sitter, and then fed into the MPC.
With an augmented state estimate, the MPC
solves a receding-horizon optimal control problem online. The next predicted state $\stateOmegaB$ and $\stateApBx$ will be used to command a low-level controller.
Finally, a low-level controller transforms $\ctrlOmegaB$ and $\ctrlApx$ commands to actuate the vehicle.
\label{sec:op-constraints}
\Para{overview-08}
The proposed framework operates under two central constraints:
\begin{itemize}
    \item \textit{Wind-free condition:} The aircraft velocity is assumed equal to the airspeed, meaning no wind-induced aerodynamic disturbances are present.
    \item \textit{Coordinated flight:} The MPC maintains coordination by tracking lateral-axis references, thereby suppressing sideslip; sustained sideslip maneuvers (e.g., severe crosswind highspeed tracking) lie outside the framework's design scope.
\end{itemize}
Within these constraints, $\cx$, $\cy$, and $\cz$ denote mass-normalized aerodynamic coefficients with units of $\mathrm{m}^{-1}$. The implementation follows a lean modeling strategy that preserves only $\cz$: the axial coefficient $\cx$ acts as a matched disturbance along the thrust axis $\xb$ and is compensated by the low-level controller as in~\cite{xuwei-acc}; the lateral coefficient $\cy$ vanishes under this coordination assumption; the aerodynamic moment is neglected because the tail-sitter's inclined motor layout provides sufficient control authority~\cite{lu2024trajectory}.
The dominance of $\cz$ is confirmed by the aerodynamic stiffness analysis in Sec.~\ref{sec:sim-benchmark}; it is therefore the only parameter proactively estimated online.
Violating either constraint introduces unmodeled aerodynamic disturbances that exceed the compensation capacity of the single-parameter online estimator; the quantitative operational boundaries under wind disturbances are characterized in Sec.~\ref{sec:op-bound}.
\section{Trajectory Generation}
\label{sec:traj-planing}
\Para{planning-01}
In this section, we establish the differential flatness of the tail-sitter by combining the kinematic structure of coordinated flight with the aerodynamic dynamics described by the $\phi$-theory model. We then formulate trajectory planning as an optimization problem based on the resulting flatness map.
\subsection{Differential Flatness}
\label{sec:differential-flatness}
\Para{planning-02}
Differential flatness allows state and control-input constraints to be handled directly during trajectory planning without explicitly integrating the system dynamics. An algebraic flatness map therefore facilitates efficient trajectory generation~\cite{minco2022}. Our derivation combines two complementary descriptions: the coordinated-flight model establishes the kinematic structure, while the $\phi$-theory model provides the aerodynamic force relations that close the dynamics-dependent part of the map.
\subsubsection{Coordinated Flight Model}
\label{sec:coordinated-flight-kinematic-model-df}
\Para{planning-03}
When in coordinated flight, the sideslip angle $\beta = \operatorname{asin} (\Bvy/V)$ of the vehicle is zero, which implies that $\Bvy = 0$. This means that the axis $\xs$ is aligned with $\stateV$.
Given the absence of aerodynamic side force in coordinated flight (i.e., $\stateAeroForceBy = \stateAeroForceSy = 0$), the second term of $\stateApS$ vanishes, i.e., $\stateApSy = 0$.
\Para{planning-04}
Under this assumption, Hauser and Hindman~\cite{df_co} prove that if $\bm{p}$ is chosen as the flat output, $\stateApS$, $\stateApDotS$ and $\stateOmegaS$ can be expressed in terms of $\bm{p}$ and its time derivatives up to the third order.
This differential flatness map must satisfy the following condition:
\begin{equation}
    \label{eq:domain-flatness}
    \left \Vert \left( \mathbf{I} - (\stateV \stateVtranspose)/V^2 \right) \stateAp \right \Vert
    \in (0, +\infty)
    .
\end{equation}
The axes of the stability frame $\mathcal{S}$ are given by
\begin{equation}
    \label{eq:stablity-frame-df}
    \xs = \frac{\stateV}{\norm{\stateV}}, \
    \zs = -\frac{\left( \mathbf{I} - \xs \, \xstranpose \right) \stateAp }{\norm{\left( \mathbf{I} - \xs \, \xstranpose \right) \stateAp }} ,\
     \ys = \zs \times \xs.\\
\end{equation}
The remaining two components of the proper acceleration $\stateApS$ are then determined as
\begin{equation}
    \label{eq:df-ap-transform}
    \stateApSx = \xs \cdot \stateAp, \quad
    \stateApSz= \zs \cdot \stateAp.
\end{equation}
According to \cite{df_co}, the jerk $\bm{j}$ of the vehicle is related to the stability-frame angular velocity $\stateOmegaS$ and the derivative of $\stateApS$, denoted by $\stateApDotS$.
The $\stateOmegaS$ is given by
\begin{subequations}
    \label{eq:df-omega-s-transform}
    \begin{align}
        \label{eq:df-omega-s-x}
        \stateOmegaSx  &= ( \stateApSx\!\stateOmegaSz - \ys \cdot \stateJ ) / \stateApSz, \\
        \label{eq:df-omega-s-y}
        \stateOmegaSy &= -(\zs \cdot \stateA) / V, \\
        \label{eq:df-omega-s-z}
        \stateOmegaSz &= (\ys \cdot \stateG) / V.
    \end{align}
\end{subequations}
Since the $\stateApDotSy = 0$, the $\stateApDotS$ is given by
\begin{equation}
    \label{eq:df-ap-S-dot}
    \stateApDotSx = \xs \cdot \stateJ - \stateOmegaSy\!\stateApSz,
    \stateApDotSz = \zs \cdot \stateJ + \stateOmegaSy\!\stateApSx.
\end{equation}
\subsubsection{$\phi$-theory Model}
\begin{figure}[tp]
    \centering
    \includegraphics[width=\linewidth]{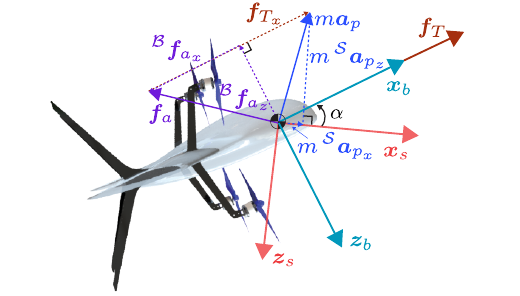}
    \caption{Geometric relationship among forces during coordinated flight, viewed in the longitudinal plane. Given a $\stateAp$ and an aerodynamic model $\stateAeroForceB(\alpha)$, states $\alpha$ and $\stateThrustForceBx$ can be determined.
    }
    \label{fig:alpha-determination}
\end{figure}
\Para{planning-05}
The dynamics-dependent part of the flatness map requires the AoA $\alpha$ to be determined from an aerodynamic model valid across the full flight envelope.
As illustrated in Fig.~\ref{fig:alpha-determination}, decomposing \eqref{eq:ap-kinematics} into the longitudinal plane, which is perpendicular to the axis $\stateYb$ and $\ys$, yields:
\begin{subequations}
    \label{eq:alpha-requirement}
    \begin{align}
    \label{eq:alpha-cal-body-frame-x}
        \stateApSx \cos \alpha
    - \stateApSz \sin \alpha
    &= \stateApBx
    =
 \left( \stateThrustForceBx + \stateAeroForceBx (\alpha)\right) / m,
    \\
    \label{eq:alpha-cal-body-frame-z}
    \stateApSz \cos \alpha+
    \stateApSx \sin \alpha
    &=\stateApBz
    =
 \stateAeroForceBz(\alpha)  / m .
    \end{align}
\end{subequations}
For complex aerodynamic maps $\stateAeroForceBx(\alpha)$ and $\stateAeroForceBz(\alpha)$,
a closed-form solution for $\alpha$ in the equations \eqref{eq:alpha-requirement} is generally unavailable \cite{eth-flywing, lu2024trajectory}.
Nevertheless, when modeled using the $\phi$-theory \cite{phi2019}, the aerodynamic force $\Bfa$ is expressed as
\begin{equation}
    \label{eq:phi-model}
    \Bfa
    =
    \begin{bmatrix}
        \Bfax \\
        \Bfay \\
        \Bfaz
    \end{bmatrix}
    =
    \begin{bmatrix}
        \phi_{11} & 0 & \phi_{13} \\
        0 & \phi_{22} & 0 \\
        \phi_{31} & 0 & \phi_{33}
    \end{bmatrix}
    \begin{bmatrix}
        \cos \alpha \\
        0 \\
        \sin \alpha
    \end{bmatrix}
    m V^2.
\end{equation}
Each $\phi_{ij}$ is a mass-normalized aerodynamic coefficient with units of $\mathrm{m}^{-1}$; the factor $m$ converts the modeled aerodynamic acceleration to force.
By substituting the third row of \eqref{eq:phi-model} into \eqref{eq:alpha-cal-body-frame-z},
the AoA $\alpha$ can be analytically solved as
\begin{equation}
\label{eq:alpha-cal-phi}
\alpha = \operatorname{atan2}\left(
\phi_{31} V^2 - \stateApSz ,
\stateApSx - \phi_{33} V^2
\right).
\end{equation}
Once $\alpha$ is solved, the rotation matrix $\RB$ can be determined from $\RS$  using \eqref{eq:transformtion-S-B}.
By substituting the first row of \eqref{eq:phi-model} into \eqref{eq:alpha-cal-body-frame-x} and rearranging terms, we derive
\begin{equation}
\label{eq:thrust-acceleration-determination}
a_T = \left(
\phi_{11} V^2 + \stateApSx
\right) \cos \alpha +
\left(
\phi_{13} V^2 - \stateApSz
\right)
\sin \alpha,
\end{equation}
where $a_T$ represents the thrust acceleration. It is assumed to be completely aligned with the axis $\xb$.
\Para{planning-06}
Up to this point, the attitude $\stateR$ and thrust acceleration $\bm{a}_T$ can be analytically determined. Furthermore, one can derive the body-frame angular velocity $\stateOmegaB$ by incorporating the jerk $\bm{j}$.
Taking the derivative of \eqref{eq:transformtion-S-B} yields:
\begin{equation}
\label{eq:omega-b-omega-s}
\begin{aligned}
    \xee{\stateOmegaB}
    = \xee{\BRS \stateOmegaS} + \xee{ \dotAlpha \, \mathbf{e}_y}.
\end{aligned}
\end{equation}
Since the map $\xee{\cdot}$ is linear and bijective, \eqref{eq:omega-b-omega-s} uniquely gives $\stateOmegaB$ as follows:
\begin{subequations}
    \label{eq:omega-b}
    \begin{align}
        \label{eq:omega-b-x}
        \stateOmegaBx &= \prescript{\mathcal{S}}{}{\bm{\Omega}_{s_x}} \cos \alpha
        -
        \prescript{\mathcal{S}}{}{\bm{\Omega}_{s_z}} \sin \alpha, \\
        \label{eq:omega-b-y}
        \stateOmegaBy &= \prescript{\mathcal{S}}{}{\bm{\Omega}_{s_y}} + \dotAlpha,\\
        \label{eq:omega-b-z}
        \stateOmegaBz &= \prescript{\mathcal{S}}{}{\bm{\Omega}_{s_x}} \sin \alpha
        +
        \prescript{\mathcal{S}}{}{\bm{\Omega}_{s_z}} \cos \alpha,
    \end{align}
\end{subequations}
where $\dotAlpha$ is the time derivative of $\alpha$.
What remains to fully determine $\stateOmegaB$ is to determine $\dotAlpha$.
By taking the derivative of \eqref{eq:alpha-cal-phi}, we can obtain
\begin{equation}
    \label{eq:alpha-dot-phi}
    \dotAlpha =
    \frac{
\phi_{31} \gamma_1 + \phi_{33} \gamma_2 + \left(\stateApSz \!\stateApDotSx - \stateApSx\!\stateApDotSz\right)
    }{
        \left( \stateApSx - \phi_{33} V^2 \right)^2 +
        \left( \phi_{31} V^2 - \stateApSz \right)^2
    },
\end{equation}
where $\gamma_1$ and $\gamma_2$ are respectively written as
\begin{equation}
    \label{eq:gamma-auxilary}
    \begin{aligned}
        \gamma_1 &= 2 V \stateApSx\!\TanAcc - V^2 \stateApDotSx, \\
        \gamma_2 &= -2 V \stateApSz \TanAcc + V^2 \stateApDotSz.
    \end{aligned}
\end{equation}
\Para{planning-07}
For now, we have established the flatness map that is necessary to formulate the trajectory optimization problem. At any time $\tau$ along a trajectory, we can define the differential flatness forward propagation $\dfForward$ as follows:
 \begin{equation}
        \label{eq:df-forward}
        \left
        (\TanAcc(\tau), \stateOmegaB(\tau)
        \right)
        =
        \dfForward(\bm{p}^{(1:3)}(\tau)).
\end{equation}
\Para{planning-08}
To establish an aerodynamic prior-free trajectory planner, the $\phi$-theory model is restricted in the $\zb$ axis, and defined solely by
$\phi_{33} = 0.5\rho S \CDflat/m$,
where $\rho$ is the air density; $S$ is the wing area.
The actual planning implementation retains only $\phi_{33}$, which is equivalent to ignoring $\cx$ in the flatness mapping.
\subsection{Spatial-temporal Trajectory Generation}
\label{sec:traj-optimization}
We parameterize the trajectory using the state-of-the-art polynomial splines trajectory representation $\MINCO$\cite{minco2022}.
The coefficients of $\MINCO$ are parameterized by intermediate waypoints $\mathbf{Q}$ and time allocation $\bm{T}$, given initial position vector $\bm{p}_0$, terminal position vector $\bm{p}_f$. The expression is:
\begin{equation}
    \label{eq:minco-class}
    \begin{aligned}
        \MINCO = & \left\{p(t):[0, T] \rightarrow \mathbb{R}^{3} \mid \mathbf{C}=\mathbf{C}(\mathbf{Q}, \bm{T};\bm{p}_0, \bm{p}_f),\right. \nonumber \\
        & \left.\mathbf{Q} \in \mathbb{R}^{3\times(M-1)}, \bm{T} \in \mathbb{R}_{++}^{M}\right\}.
    \end{aligned}
\end{equation}
where $M$ represents the number of the piecewise polynomial trajectory segments. The duration of each segment is denoted by $t_i$ and the endpoint of each segment (except the last segment), referred to as an intermediate waypoint, is represented by $\bm{q}_1, \dots, \bm{q}_{M-1}$, respectively.
The intermediate waypoints, denoted by $\mathbf{Q} = (\bm{q}_1, \dots, \bm{q}_{M-1})$, and the time allocation, denoted by $\bm{T} = (t_1, \dots, t_M)$ are decision variables during trajectory optimization. The coefficient collection $\mathbf{C} = (\bm{c}_1, \bm{c}_2, \dots, \bm{c}_M)$, where each $\bm{c}_i$ represents the coefficient of the $i$-th trajectory segment.
\Para{planning-09}
We employ 7th-order polynomial splines within the $\MINCO$ trajectory representation to ensure $C^3$ continuity across the entire trajectory~\cite{minco2022}, where the polynomial basis is $\beta(s) = (1, s, \dots, s^7)$.
Let $\tau_0=0$ and $\tau_i=\sum_{k=1}^{i}t_k$ denote the cumulative endpoint time of segment $i$, for $i=1,\dots,M$, so that $\tau_M=T_f$.
The complete trajectory is then evaluated piecewise as
\begin{equation}
    \label{eq:traj-segment}
    \bm{p}(\tau) = \bm{c}_i \cdot \beta(\tau - \tau_{i-1}), \quad \tau \in [\tau_{i-1}, \tau_i].
\end{equation}
Building upon $\dfForward$ and $\MINCO$, we formulate the trajectory optimization problem as
\begin{subequations}
\label{eq:traj-optimization}
\begin{align}
    \min _{\mathbf{Q}, \bm{T}} \hspace{.5em}
    &\int_{0}^{T_{f}}\left\|\bm{p}^{(4)}(\tau)\right\|^{2} \operatorname{d}\!\tau+ \rho T_{f},
    \label{eq:original-constr}
    \\
\operatorname{s.t.}\hspace{.5em} &\bm{p}(t)=\MINCO(\trajDecWaypoints, \trajDecTimeAllocation; \bm{p}_0, \bm{p}_f),
\quad T_{f}=\sum_{i=1}^{M} t_{i},
\label{eq:coef-contr}
\\
&\bm{q}_{i} \in \mathcal{Q}_i, \quad i=1,\dots, M-1,
\label{eq:corridor-constr}
\\
&\norm{\bm{v}(t)} \leq \plnMaxVel, \label{eq:vel-constr}\\
&-\!\plnMinTanAcc
\leq \TanAcc(t)
\leq \plnMaxTanAcc,
\label{eq:feas-contr-at}
\\
&-\!\plnMaxOmega \leq \stateOmegaB(t)
\leq \plnMaxOmega,
\label{eq:feas-contr-omega}
\end{align}
\end{subequations}
where $\rho>0$ is the penalty weight for the total flight time $T_f$, $\mathcal{Q}_1 \times \mathcal{Q}_2 \times \cdots \times \mathcal{Q}_{M-1}$ is the allowable region for the intermediate waypoints $\mathbf{Q}$ to freely move, $\plnMaxVel$ denotes the maximum velocity, $\plnMaxOmega$ denotes the maximum body-frame angular velocity, and $\plnMaxTanAcc$ and $\plnMinTanAcc$ represent the maximum and minimum tangential acceleration, respectively.
The constraints \eqref{eq:vel-constr}--\eqref{eq:feas-contr-omega} are set for dynamically feasible trajectory planning. Notably, the forward and backward transition times are directly regulated by the tangential acceleration limits, $\plnMaxTanAcc$ and $\plnMinTanAcc$, respectively.
\begin{algorithm}[t]
\caption{Aerodynamic Prior-Free Trajectory Planning}
\label{alg:traj-pln}
\begin{algorithmic}[1]
\State
\Givens{initial intermediate waypoints $\trajDecWaypoints^0$, initial time allocation $\trajDecTimeAllocation^0$, the minimum rate of decrease of the cost function $\delta $
and translation-invariant shape $\mathcal{G}$ for each $\mathcal{Q}_i$, where $i=1, \cdots, M-1$.
}
\State $k \gets 0, \mathbf{B} \gets \mathbf{I}, \norm{\nabla \costPlanning} \gets \infty$
\While{$\norm{\nabla \costPlanning} > \delta$}
\State $\bm{p}^{(0:4)}(t) \gets \MINCO(\trajDecWaypoints^k, \trajDecTimeAllocation^k)$
        \State $\TanAcc(t), \stateOmegaB(t) \gets$ $\dfForward(\bm{p}^{(1:3)}(t))$
        \State $\nabla \costConstr \gets$ $\dfBackward\left(\TanAcc(t), \stateOmegaB(t), \bm{p}^{(1)}(t)\right)$
        \State $\nabla \costOriginal
        \gets \textsc{Gcopter}(\bm{p}^{(0:4)}(t), \mathcal{G})
        $ \cite{minco2022}
        \State $\nabla \costPlanning \gets
        \nabla \costOriginal + \nabla \costConstr
        $
        \State $\mathbf{d}^k \gets -\mathbf{B}^k \, \nabla \costPlanning$
        \State $\trajDecWaypoints^{k+1},\trajDecTimeAllocation^{k+1}  \gets
        \textsc{Line Search}(\trajDecWaypoints^k, \trajDecTimeAllocation^k, \mathbf{d}^k)$ \cite{lewis-line}
        \State $\mathbf{B}^{k+1} \gets$ L-BFGS \cite{lbfgs}
        \State $k \gets k + 1$
    \EndWhile
    \State \Return $\MINCO(\mathbf{Q}^{*}, \bm{T}^{*})$
\end{algorithmic}
\end{algorithm}
\Para{planning-10}
The optimization problem \eqref{eq:traj-optimization} aims to find a dynamically feasible trajectory, with the decision variables being the intermediate waypoints $\mathbf{Q}$ and the time allocation $\bm{T}$.
This formulation is based on GCOPTER \cite{minco2022}, a versatile trajectory planning framework that excels at managing constraints on intermediate waypoints within flight corridors and total time allocation.
By leveraging the $\MINCO$ representation, GCOPTER transforms these complex spatial and temporal trajectory planning requirements into an unconstrained nonlinear programming problem (NLP).
Consequently, constraints \eqref{eq:coef-contr} and \eqref{eq:corridor-constr} are inherently satisfied within this formulation, while the constraint violation terms for \eqref{eq:vel-constr}--\eqref{eq:feas-contr-omega} are incorporated into the GCOPTER-transformed cost function using a quadratic exterior penalty method.
The cost function of the unconstrained NLP, denoted as $\costPlanning$, consists of two components. The first component, $\costOriginal$, includes the original cost function \eqref{eq:original-constr} and the constraint violation penalty terms for  \eqref{eq:coef-contr} and \eqref{eq:corridor-constr}.
The gradient of the corresponding terms for $\costOriginal$  is properly addressed in \cite{minco2022}.
The second component, $\costConstr$, represents the penalty terms for violating constraints
in \eqref{eq:vel-constr}-\eqref{eq:feas-contr-omega}.
The crucial part of this NLP lies in the gradient backpropagation associated with the penalty terms for the constraint violations in \eqref{eq:feas-contr-at} and \eqref{eq:feas-contr-omega}.
The corresponding gradient terms for \eqref{eq:feas-contr-at} and \eqref{eq:feas-contr-omega} are composed of two parts as follows:
\begin{equation}
        \frac{\partial \left(\TanAcc(t), \stateOmegaB(t) \right)}
        {\partial \,(\mathbf{Q}, \bm{T})}
        = \frac{\partial \left(\TanAcc(t), \stateOmegaB(t)\right)}{\partial \,\bm{p}^{(1:3)}}
        \frac
        {\partial \,\bm{p}^{(1:3)}}
        {\partial \,(\mathbf{Q}, \bm{T})}.
\end{equation}
The second part, $
        {\partial \,\bm{p}^{(1:3)}}
        /{\partial \,(\mathbf{Q}, \bm{T})}$, has been resolved in \cite{minco2022}.
To avoid the cumbersome and error-prone symbolic derivation of the first part $\partial \left(\TanAcc(t), \stateOmegaB(t)\right) / {\partial \,\bm{p}^{(1:3)}}$, we obtain this
using the forward mode automatic differentiation method, as described in \cite{autodiff}.
The constraint violations from \eqref{eq:vel-constr} to \eqref{eq:feas-contr-omega} are checked
at uniformly spaced samples along the entire trajectory. When a sample violates these constraints, the corresponding penalty gradient terms are backpropagated to the decision variables using the chain rule. This process is referred to as differential flatness backpropagation, denoted as $\dfBackward$.
To achieve an appropriate gradient descent step and direction, we further integrate the gradient terms of the NLP within a line search framework that satisfies the weak Wolfe condition \cite{lewis-line}, alongside the L-BFGS quasi-Newton gradient method \cite{lbfgs}. The whole aerodynamic prior-free trajectory planning algorithm is seen in Alg.~\ref{alg:traj-pln}.
\section{Trajectory tracking control}
\label{sec:traj-tracking-controller}
\Para{control-01}
Under the modeling strategy established in Sec.~\ref{sec:op-constraints}, we develop an innovative aerodynamic prior-free MPC. By incorporating an online estimated aerodynamic parameter, the MPC achieves high-accuracy tracking of coordinated flight trajectories.
\subsection{Prediction Model}
\label{sec:modelDesign}
\Para{control-03}
Using the mass-normalized aerodynamic coefficients $\cx$, $\cy$, and $\cz$ defined in Sec.~\ref{sec:op-constraints}, we set $\cy = 0$, $\cx = 0$, and $\aeroMoment = \bm{0}$. The aerodynamic effects in the MPC are thus modeled as
\begin{equation}
    \label{eq:nmpc-aerodynamics}
    \stateAeroForce = m \cz V^2 \zb, \quad \aeroMoment = \bm{0},
\end{equation}
\Para{control-04}
In the aerodynamic model described by \eqref{eq:nmpc-aerodynamics}, the effects are assumed to depend solely on the parameter $\cz$.
Since the model only needs to remain locally valid over the MPC's short prediction horizon, we can effectively implement adaptive control by treating this parameter as a time-varying state. Specifically, we augment the nominal state $\bm{x}$ with the aerodynamic parameter $\cz$ for online estimation, defining the $\cz$ dynamics as
\begin{equation}
    \label{eq:cz-dynamics}
    \DotCz = \kappa \dotAlpha,
\end{equation}
where $\kappa$ is a fixed gain tuned for performance. This linear model \eqref{eq:cz-dynamics} allows the MPC to know that pitch down typically ($\stateOmegaBy < 0$) causes the magnitude of $\stateAeroForceBz$ to decrease, and pitch up typically ($\stateOmegaBy > 0$) causes the magnitude of $\stateAeroForceBz$ to increase.
The $\dotAlpha$ can be obtained by rewriting \eqref{eq:omega-b-y} and expanding the expression of \eqref{eq:df-omega-s-y} using \eqref{eq:expansion-axes} into
the frame $\mathcal{B}$
as follows:
\begin{equation}
    \label{eq:nmpc-alpha-dot}
    \dotAlpha = \stateOmegaBy + \frac{\zs \cdot \stateA}{V}
    = \stateOmegaBy
    + \frac{\aBz \cos \alpha -\aBx \sin \alpha}{V}.
\end{equation}
Note that the $\dotAlpha$ calculated from \eqref{eq:nmpc-alpha-dot} depends entirely on the longitudinal motion of the tail-sitter, and is independent of the lateral motion that is characterized by $\stateOmegaBx$, $\stateOmegaBz$, and $\aBy$.
\Para{control-05}
There are four motors in the tail-sitter, and the thrust force $\stateFT$ and thrust moment are modeled as
\begin{equation}
    \label{eq:thrust-model}
    \begin{bmatrix}
        \stateFTBx \\
        \stateMomentTBx \\
        \stateMomentTBy \\
        \stateMomentTBz
    \end{bmatrix}
    =
    \begin{bmatrix}
        1 & 1 & 1 & 1 \\
        c_M, & c_M & -c_M & -c_M \\
        d_z & -d_z & d_z & -d_z \\
        -d_y & -d_y & d_y & d_y
    \end{bmatrix}
    \begin{bmatrix}
        f_{T_1} \\
        f_{T_2} \\
        f_{T_3} \\
        f_{T_4}
    \end{bmatrix},
\end{equation}
where $d_y$ and $d_z$ are the motor displacements, $c_M$ is the propeller torque constant.
The control thrust force $\fc$ is defined to be the four actuators' forces vector, i.e., $\fc = [f_{T_1}, f_{T_2}, f_{T_3}, f_{T_4}]^{\text{T}} \in \RR^4$.
We further augment the nominal state $\bm{x}$ with $\fc$ and the MPC control input $\bm{u}$ is defined as follows:
\begin{equation}
\label{eq:nmpc-fc-dynamcis}
\bm{u} = [\dot{f}_{T_1}, \dot{f}_{T_2}, \dot{f}_{T_3}, \dot{f}_{T_4}]^{\text{T}} \in \RR^4
\end{equation}
This augmentation allows us to constrain the control force and moment generated by the actuators.
\Para{control-06}
Overall, in linear acceleration prediction, we simplify \eqref{eq:ap-kinematics} as
\begin{equation}
    \label{eq:nmpc-ap-simple-modeling}
    \stateAp = a_{T} \, \stateXb + \cz V^2 \, \stateZb.
\end{equation}
Here, $a_T=\stateFTBx/m$ is the thrust acceleration, so both terms in \eqref{eq:nmpc-ap-simple-modeling} have units of acceleration.
In angular acceleration prediction, we ignore the aerodynamic moment and simplify \eqref{eq:rotational-kinematics-Omega} as
\begin{equation}
    \label{eq:nmpc-moment-simple-modeling}
    \stateDotOmegaB = \mathbf{J}^{-1}
    \left(
    \stateMomentT - \xee{\stateOmegaB} \mathbf{J} \stateOmegaB
    \right).
\end{equation}
The augmented state is defined as $\bm{s} = \left(
    \bm{x}, \cz , \fc
\right)$. By combining \eqref{eq:translational-kinematics}, \eqref{eq:rotational-kinematics-R}, \eqref{eq:cz-dynamics}, \eqref{eq:nmpc-alpha-dot}, \eqref{eq:thrust-model}, \eqref{eq:nmpc-fc-dynamcis}, \eqref{eq:nmpc-ap-simple-modeling} and \eqref{eq:nmpc-moment-simple-modeling}, the state transfer function can be expressed compactly as
\begin{equation}
    \label{eq:nmpc-prediction-model}
    \dot{\bm{s}} = f_{\text{MPC}}(\bm{s}, \bm{u}).
\end{equation}
The proposed prediction model not only retains the most significant aerodynamic effect in a controller's perspective but also is locally valid within the full flight envelope.
\subsection{MPC Optimization Problem Formulation}
\label{sec:mpc-cost}
The prediction model \eqref{eq:nmpc-prediction-model} is designed for and highly effective in coordinated flight, but loses validity under significant sideslip aerodynamics. To address this limitation, we propose a coordinated flight strategy that exploits the flatness in Sec.~\ref{sec:differential-flatness}, and is deeply integrated with the MPC.
\Para{control-07}
Given a $C^3$ continuous reference position trajectory $\refposTraj$ by the coordinated-flight planner in Sec.~\ref{sec:traj-planing}, the trajectories $\refvelTraj$, $\refaccTraj$, and $\refybTraj$ can be obtained using the flatness in Sec.~\ref{sec:differential-flatness}. We first assume the $\refposTraj$ is accurately tracked, in the prediction horizon $[t_0, t_f]$ of the MPC, we have
\begin{equation}
    \label{eq:vel-accel-assumption}
    \stateV(\tau) = \refvel(\tau), \quad \stateAp(\tau) = \refacc(\tau), \quad \tau \in [t_0, t_f].
\end{equation}
Therefore, accurately tracking the trajectory $\refposTraj$ implies
\begin{equation}
    \label{eq:yb-assumption}
    \norm{
        \stateYb(\stateV(\tau), \stateAp(\tau)) - \refyb( \refvel(\tau),  \refacc(\tau))
    } = 0, \quad \tau \in [t_0, t_f].
\end{equation}
This hard constraint \eqref{eq:yb-assumption} is softened into the cost function of the MPC as follows:
\begin{equation}
    \label{eq:cost-nmpc}
    \begin{aligned}
        \costMPC =
        \int_{t_0}^{t_f}
        \Big(
        &\underbrace{\left\|\refpos-\stateP\right\|_{\mathbf{Q}_{\stateP}}^2+
        \left\|\refvel-\stateV\right\|_{\mathbf{Q}_{\stateV}}^2 }_{\text{trajectory tracking cost}}\\
        &+\underbrace{\left\|\refyb-\yb\right\|_{\mathbf{Q}_{\yb}}^2}_{
            \text{coordinated flight cost}}
         +\underbrace{
            \left\|\bm{u}\right\|_{\mathbf{R}_{\bm{u}}}^2}_{
                \text{control effort cost}}
        \Big) \mathrm{d}\,\tau.
    \end{aligned}
\end{equation}
The $\mathbf{Q}_{\stateP}, \mathbf{Q}_{\stateV}, \mathbf{Q}_{\yb}, \mathbf{R}_{\bm{u}}$ are all positive-definite diagonal matrices.
The control effort cost in \eqref{eq:cost-nmpc} is used to penalize the rate of control thrust force $\fc$ to ensure dynamic feasibility.
According to \eqref{eq:thrust-model} and \eqref{eq:nmpc-moment-simple-modeling},
the dynamics of $\fc$ dominates those of $\stateOmegaB$ and $a_T$. Consequently, the control effort cost inherently and consistently penalizes high-frequency components in $\ctrlat$ and $\ctrlOmegaB$, leading to a steady control behavior.
\Para{control-08}
To ensure dynamical feasibility and controllability of the low-level controller, we further impose the following constraints on the augmented state and control input within the MPC prediction horizon:
\begin{subequations}\label{eq:mpc-constraints}
\begin{align}
    & -\!\overline{\stateOmegaB} \leq \prescript{\mathcal{B}}{}{\mathbf{\Omega}_{b}} \leq \overline{\stateOmegaB}
\label{eq:nlp-cont-bd-omega}
    \\
    & 0 \leq {\bm{f}}_{c} \leq \overline{\fc} \label{eq:nlp-cont-bd-T}
    \\
& -\!\overline{\bm{u}}
 \leq \bm{u}  \leq \overline{\bm{u}}
\label{eq:nlp-cont-bd-u}
\end{align}
\end{subequations}
The constraints \eqref{eq:nlp-cont-bd-omega} and \eqref{eq:nlp-cont-bd-T} are dictated by the physical actuator limits. Meanwhile, the constraint \eqref{eq:nlp-cont-bd-u} and the control effort cost in \eqref{eq:cost-nmpc} are jointly tuned to constrain the bandwidth of the commands sent to the low-level controller.
\Para{control-09}
The continuous-time MPC problem is transcribed into a NLP and solved via sequential quadratic programming (SQP). The resulting QP subproblems are solved in real time by HPIPM~\cite{hpipm}, the state-of-the-art interior-point method QP solver tailored for MPC problems. The implementation is facilitated by acados \cite{acados} and CasADi \cite{casadi}.
The MPC parameters used in both simulation and field experiments are summarized in Table~\ref{tab:mpc-params}, except $\mathbf{Q}_{\bm{p}_z}$ set to $5$ in field experiments due to noisy height estimates. Notably, the weight $\mathbf{Q}_{\bm{y}_b}$ is set to be highest, consistent with our coordinated flight assumption.
\begin{table}[t]
    \centering
    \caption{MPC Configuration Parameters}
    \label{tab:mpc-params}
    \begin{tabular}{ll}
        \toprule
        Parameter & Value \\
        \hline
        State weight $(\mathbf{Q}_p; \mathbf{Q}_v; \mathbf{Q}_{\yb})$ & $\mathrm{diag}(15, 15, 10; 1, 1, 1; 52.5, 52.5, 126)$ \\
        Control effort weight $\mathbf{R}$ & $\mathrm{diag}(0.03, 0.03, 0.03, 0.03)$ \\
        \hline
        Thrust rate upper bound $\overline{\bm{u}}$ & $\mathrm{diag}(5, 5, 5, 5)\, \text{m/s}^3$ (mass-normalized)\\
        Bodyrate upper bound $\overline{\stateOmegaB}$ & $\mathrm{diag}(1.5, 2.5, 1.5)$\,rad/s \\
        \bottomrule
    \end{tabular}
\end{table}
\subsection{Augmented State Estimation}
\label{sec:init-val}
\Para{control-10}
As part of the standard procedure, the nominal state components of the augmented state---$\stateP$, $\stateV$, $\stateR$, and $\stateOmegaB$---are typically obtained using a standard state estimator such as an extended Kalman filter (EKF).
In this section, we will focus on the estimation of the augmented components $\fc$ and $\stateAeroCoefz$.
\Para{control-12}
Rather than estimating $\fc$ from motor-speed measurements, which demands extra hardware (e.g., Bidirectional DShot or optical encoders~\cite{mit_estimation}) and a pre-identified propeller model, we note that the faster inner loop handles $\fc$ transients and that $\bm{u}$ is already rate-limited in~\eqref{eq:nlp-cont-bd-u}; we therefore simply initialize $\fc$ to the last MPC result.
To estimate $\cz$, we subtract the thrust acceleration $\stateFT/m$ from the proper acceleration, project the result onto $\zb$, and divide by $V^2$, yielding:
\begin{equation}
    \label{eq:cz-estimaion-theory}
    \cz = \frac{\left(\stateAp - \frac{\stateFT}{m}\right) \cdot \zb}{V^2}
        = \frac{\stateApBz}{V^2}.
\end{equation}
The second equality follows from $\stateFT \cdot \zb = 0$ because the thrust force is aligned with $\xb$.
Following \eqref{eq:cz-estimaion-theory}, the estimated
$\cz$ is computed from filtered acceleration measurements using a small denominator offset and clamping to prevent singularities and over-estimation.
\section{Experiments}
\label{sec:exp}
\Para{experiments-01}
In this section, we evaluate the proposed aerodynamic prior-free flight framework through a series of simulation and field experiments.
In simulation, we evaluate the efficacy of the proposed approach by benchmarking it against a diverse set of state-of-the-art controllers. We also investigate the framework's operational boundaries under wind disturbances.
Subsequently, field experiments validate the complete pipeline through standard tracking tasks and aggressive fast transitions. Notably, the vehicle maintains precise trajectory execution without aerodynamic priors, even under aggressive conditions with a peak velocity of 13.3\,m/s and roll and pitch angles reaching 48$^{\circ}$ and 78$^{\circ}$. For better visualization, please refer to \href{https://youtu.be/H4pDQo8YX98}{https://youtu.be/H4pDQo8YX98}.
\begin{figure}[t]
    \centering\includegraphics[width=\linewidth]{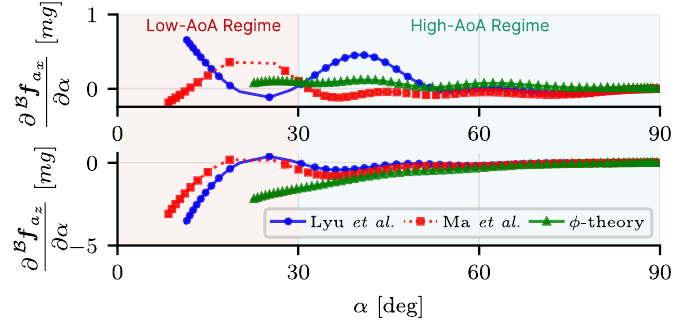}
    \caption{Longitudinal aerodynamic stiffness characteristics evaluated at trim conditions spanning flight velocities from 1 to 14\,m/s. The results indicate that the low-AoA regime is highly sensitive to attitude perturbations, while the high-AoA regime is significantly less sensitive. The two regimes are shaded qualitatively for illustrative purposes.}
    \label{fig:aeros}
\end{figure}
\subsection{Benchmark under Various Aerodynamic Conditions}
\label{sec:sim-benchmark}
\Para{experiments-02}
\begin{figure*}
    \centering
    \includegraphics[width=1.0\textwidth]{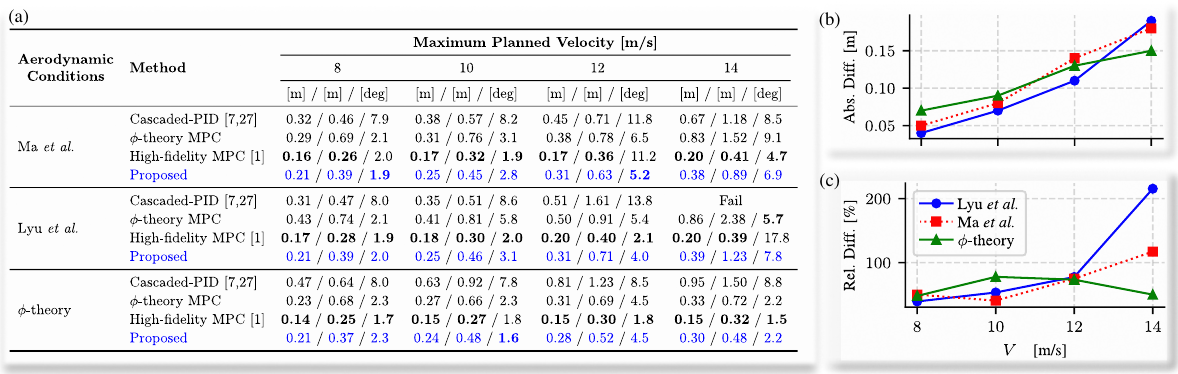}
    \caption{Quantitative comparison of tracking performance over the tested flight regimes. (a) Position RMSE, MaxAE, and maximum sideslip angle. During transition phases, the sideslip angle may exceed reported values as the coordinated flight constraint is enforced via an exterior penalty method. The table compares the proposed method against three baseline controllers: the Cascaded-PID (reactive aerodynamic prior-based), the $\phi$-theory MPC (lower bound), and the high-fidelity MPC (upper bound). (b) Absolute RMSE difference between the proposed method and the high-fidelity MPC upper bound. (c) Relative MaxAE difference between the proposed method and the high-fidelity MPC upper bound, quantifying worst-case performance degradation. Note that MaxAE for both the proposed method and the upper bound occurs within the transition phase, where nonlinear aerodynamic effects are most severe.}
    \label{fig:sim_tracking}
\end{figure*}
The goal of this benchmark is twofold: 1) to evaluate whether the aerodynamic prior-free framework can achieve robust performance across diverse aerodynamic conditions without prior knowledge; 2) to quantify the performance gap between the proposed method and prior-based approaches.
To achieve these objectives, simulations were performed in a high-fidelity Gazebo/PX4 co-simulation environment under wind-free conditions. Three aerodynamic models served as virtual plants to represent diverse real-world scenarios: wind-tunnel-identified models from Ma et al.~\cite{MaAero} and Lyu et al.~\cite{lyu-cas}, and an idealized $\phi$-theory model~\cite{phi2019}. As shown in Fig.~\ref{fig:aeros}, these aerodynamic models are scaled to produce comparable effects. Both $\partial\Bfaz/\partial\alpha$ and $\partial\Bfax/\partial\alpha$ are moderate in the high-AoA regime; as a result, aerodynamic modeling errors may not contribute as significantly to translational tracking error in hovering and slow-speed flight phases. Accordingly, the MPC prediction model retains only $\cz$ for two reasons. First, in the low-AoA regime where aerodynamic effects most significantly impact tracking, $\partial\Bfaz/\partial\alpha$ substantially exceeds $\partial\Bfax/\partial\alpha$, making $\cz$ estimation the dominant factor in translational accuracy. Second, the residual $\cx$-induced force acts as a matched disturbance along the fully actuated $\xb$ axis and is suppressed by the low-level controller as in~\cite{xuwei-acc}.
\Para{experiments-03}
To assess performance across these aerodynamic regimes, we designed circular trajectory tracking experiments spanning velocities from 8 to 14\,m/s. Each trajectory comprises three phases: forward transition, level flight, and backward transition.
We evaluate these experiments using position root mean square error (RMSE) and maximum absolute error (MaxAE) during transition and level-flight phases, and sideslip angle $\beta$ during level flight. These metrics indicate average accuracy, worst-case behavior, and coordinated flight quality, respectively.
The baseline controllers selected for this comparison include:
\begin{itemize}
    \item Cascaded-PID~\cite{quanquan-liftwing,df-lift-wing}, which represents a flatness-based cascaded control architecture where the attitude command is synthesized by combining the differential flatness mapping described in Sec.\ref{sec:differential-flatness} with the AoA obtained via an optimization subroutine solving \eqref{eq:alpha-cal-body-frame-z} based on high-fidelity aerodynamic models.
    \item $\phi$-theory MPC, an MPC-based baseline utilizing the same simplified aerodynamic model employed during the trajectory planning phase.
    \item High-fidelity MPC~\cite{lu2024trajectory}, which linearizes the system dynamics along a reference trajectory using a high-fidelity aerodynamic model to derive a time-varying error-state model.
\end{itemize}
\Para{experiments-04}
As illustrated in Fig.~\ref{fig:sim_tracking}(a), both the cascaded-PID controller and the $\phi$-theory MPC exhibit higher tracking errors than the other tested methods. Under 12 m/s, the $\phi$-theory MPC demonstrates performance comparable to the prior-based cascaded-PID, while maintaining crash-free operation across all test cases. Such behavior validates the simplified aerodynamic model adopted in our planning phase, establishing a reasonable lower bound for the proposed framework. However, the MaxAE of the $\phi$-theory MPC exceeds 1.5\,m under realistic aerodynamic conditions, which reveals the limitations of the $\phi$-theory model over the tested flight regimes.
\Para{experiments-05}
Overall, the high-fidelity MPC outperforms the proposed method across all metrics, confirming its role as a performance upper bound. Nevertheless, the proposed MPC consistently achieves less than 0.2\,m absolute RMSE difference from the high-fidelity MPC across all tested aerodynamic conditions, even at the maximum tested velocity of 14\,m/s (Fig.~\ref{fig:sim_tracking}(b)). This demonstrates that the proposed method serves as an effective initial controller that delivers competitive tracking accuracy without airframe-specific aerodynamic priors. However, the MaxAE degrades substantially in the high-speed regime where aerodynamic forces dominate. As shown in Fig.~\ref{fig:sim_tracking}(c), the worst-case relative MaxAE difference exceeds 100\,\% under the Ma et al.\ aerodynamic condition and 200\,\% under the Lyu et al.\ condition at 14\,m/s. These results indicate that at very high speeds, where aerodynamic sensitivity is extreme (Fig.~\ref{fig:aeros}), high-fidelity aerodynamic modeling further reduces worst-case tracking error.
\begin{figure}[htbp]
    \centering
    \includegraphics[width=\linewidth]{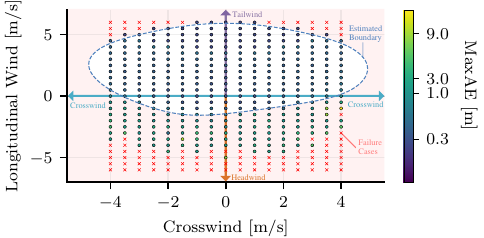}
    \caption{Wind disturbance experiment under Lyu \textit{et al.} aerodynamic condition.
    The empirical tracking-performance boundary (dashed line), estimated via a Gaussian process classifier, delineates the tested wind region where the controller maintains $\text{MaxAE} \leq 1$\,m; it is not a formal safety or reachability guarantee.
    Outside this boundary, aerodynamic force misprediction gradually decouples the predicted state from the true plant while the vehicle continues following the reference trajectory; the growing mismatch eventually prevents the optimization from converging and causes the solver to fail.
    }
    \label{fig:wind}
\end{figure}
\subsection{Operational Boundaries}
\label{sec:op-bound}
\Para{experiments-06}
To characterize the empirical tracking-performance boundary of the proposed aerodynamic prior-free flight framework, we conducted extensive wind-disturbance simulation sweeps.
In each simulation trial, the vehicle tracked a consistent full-envelope trajectory at a maximum velocity of 12\,m/s, subject to varying longitudinal and crosswind disturbances.
As shown in Fig.~\ref{fig:wind}, the interplay between the wind-free assumption and the simplified estimation~\eqref{eq:cz-estimaion-theory} produces direction-dependent empirical tracking limits.
\Para{experiments-07}
\begin{figure*}[t]
    \centering
    \includegraphics{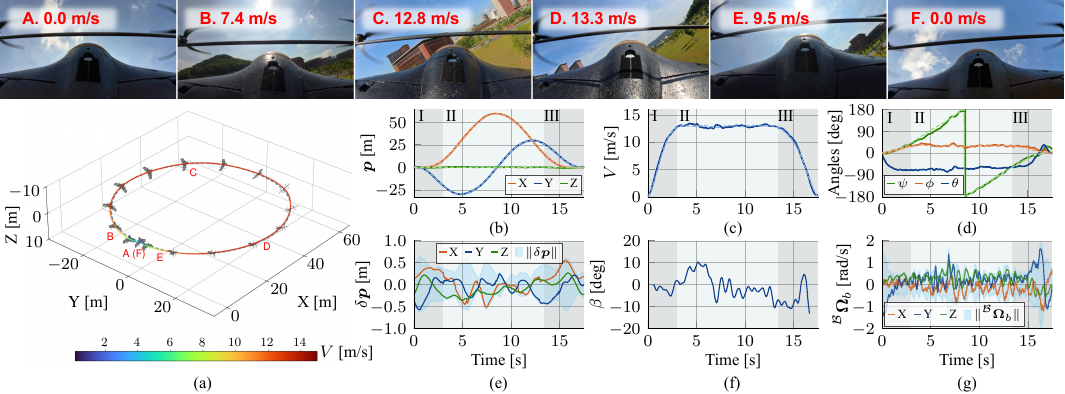}
        \caption{Flight data of the circular flight in 13\,m/s: (a) trajectory illustration; (b) position; (c) flight speed; (d) lateral axis angles ($\psi$ and $\phi$) and pitch angle; (e) position error; (f) ground-velocity-based sideslip proxy $\hat{\beta}_{g}$ (labeled $\beta$ in the plot); and (g) body-frame angular velocity. Flight phases from I to III, divided by shaded areas, indicate the forward transition, level-flight, and backward transition. }
        \label{fig:circular}
\end{figure*}
This performance asymmetry stems from how wind direction shifts the vehicle's aerodynamic operating point between the two characteristic regimes identified in Sec.~\ref{sec:sim-benchmark} (Fig.~\ref{fig:aeros}). Headwinds force the vehicle into the low-AoA regime, where high aerodynamic stiffness induces rapid force variations that challenge the fidelity of the model~\eqref{eq:nmpc-aerodynamics} and the accuracy of the simplified estimation~\eqref{eq:cz-estimaion-theory}. Tailwinds shift equilibria to the high-AoA regime, where sufficient control authority enables stable flight. Additionally, crosswinds induce sideslip, violating the MPC's coordination constraint~\eqref{eq:yb-assumption} and potentially invalidating the internal formulation for optimal coordinated trajectory tracking. Collectively, the proposed flight framework is most sensitive to combined headwind and crosswind.
\begin{figure*}[t]
    \centering
    \includegraphics{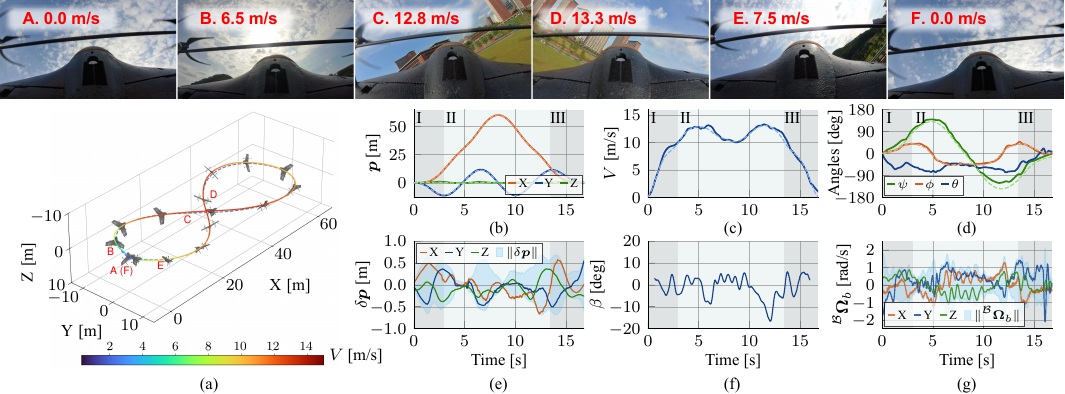}
        \caption{Flight data of the $\infty$-shaped flight in 13\,m/s: (a) trajectory illustration; (b) position; (c) flight speed; (d) lateral axis angles ($\psi$ and $\phi$) and pitch angle; (e) position error; (f) ground-velocity-based sideslip proxy $\hat{\beta}_{g}$ (labeled $\beta$ in the plot); and (g) body-frame angular velocity. Flight phases from I to III, divided by shaded areas, indicate the forward transition, level-flight, and backward transition.}
        \label{fig:lemniscate}
\end{figure*}
\begin{figure*}[t]
    \centering
    \includegraphics[width=\textwidth]{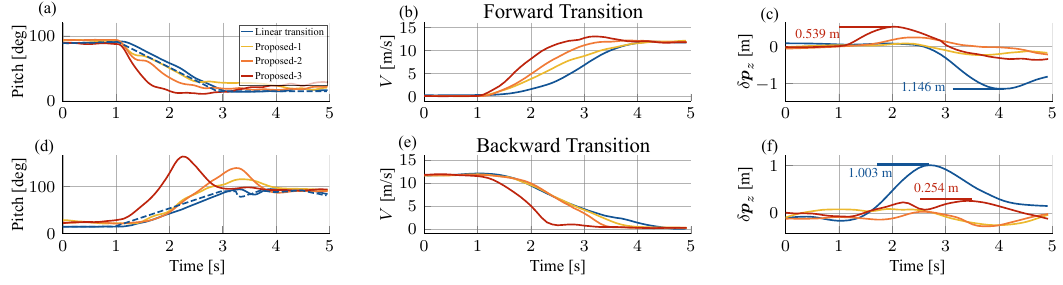}
        \caption{Comparison of rapid transition performance between the proposed method and the linear transition method~\cite{lyu-cas}. (a)--(c) show the pitch angle, flight speed, and altitude tracking error during the forward transition, and (d)--(f) for the backward transition.
        Proposed-1 to Proposed-3 correspond to the results with increasing levels of transition aggressiveness. The dashed lines in (a) and (d) indicate the reference profiles of the linear transition method; in contrast, the proposed MPC directly commands the pitch rate without a pitch angle reference.}
        \label{fig:rapid-transition}
\end{figure*}
\subsection{Field Experiments}
\subsubsection{Experimental Setup}
\Para{experiments-08}
The tail-sitter is equipped with four SZ-SPEED 2312-1000 KV motors and 9450 self-tightening propellers.
It weighs 2.0~kg and has a wingspan of 110~cm.
Moreover, it is equipped with an onboard computer Intel N100 (4 cores, 1.8--3.2 GHz, 6W TDP) and a Pixhawk 6X autopilot.
A separate Holybro H-RTK F9P Helical module is mounted to provide positioning data. The vehicle state is estimated by an EKF running on the autopilot.
Raw three-axis acceleration readings are filtered using second-order low-pass Butterworth filters with a 10 Hz cutoff to estimate $\stateAeroCoefz$.
The experiments were conducted outdoors under light-wind conditions (wind speed below 2\,m/s) without an airspeed sensor. These tests therefore evaluate practical performance under mild deviations from, rather than strict satisfaction of, the wind-free modeling assumption.
The proposed trajectory planner and controller are implemented on the N100. Trajectory planning runs offline, while the MPC operates at 50\,Hz and the low-level controller at 400\,Hz.
The MPC optimization takes an average of 2\,ms and a maximum of around 5\,ms.
The low-level loop, realized as simple PID controllers per~\cite{xuwei-acc}, tracks the reference $\stateApBx$ and $\stateOmegaB$. These commands are provided by the MPC's predicted next state. The resulting control outputs are then mapped to four DSHOT-600 motor commands via a standard quadcopter mixer.
\subsubsection{Typical Maneuvers in Field Environments}
\Para{experiments-09}
To demonstrate the trajectory tracking performance of the proposed framework, we evaluate two types of trajectories: circular and $\infty$-shaped. Both trajectories consist of three phases: forward transition, level flight, and backward transition.
Fig.~\ref{fig:circular} and Fig.~\ref{fig:lemniscate} show the circular and $\infty$-shaped trajectories, respectively, and their corresponding tracking flight data.
As illustrated in both figures, the tail-sitter first performs a forward transition (Phase I) with significant linear acceleration and a large negative $\stateOmegaBy$ to gain velocity for level flight (Phase II). After level flight, the tail-sitter executes a backward transition (Phase III) with substantial linear deceleration and a large positive $\stateOmegaBy$ to reduce velocity and achieve hovering.
\Para{experiments-10}
It is observed that the position, velocity, and lateral axis estimates closely track the reference values throughout the flight.
Because no airspeed sensor is installed, panel (f) reports a ground-velocity-based sideslip proxy $\hat{\beta}_{g}$ rather than the true aerodynamic sideslip angle $\beta$. For nearly the entire level-flight duration in both the circular and $\infty$-shaped trajectories, $|\hat{\beta}_{g}|$ remains within $10^\circ$, indicating close alignment between the ground-velocity direction and the body longitudinal plane. This ground-track alignment is maintained using the low-level control method from~\cite{xuwei-acc} without yawing torque compensation.
For the circular trajectory shown in Fig.~\ref{fig:circular}, the position RMSE and MaxAE are 0.39\,m and 0.66\,m, respectively. In contrast, the $\infty$-shaped trajectory, shown in Fig.~\ref{fig:lemniscate}, exhibits slightly higher errors, with an average of 0.42\,m and a maximum of 0.73\,m. The position tracking error $\delta \bm{p}$ throughout the entire flight is shown in Fig.~\ref{fig:circular}(e) and Fig.~\ref{fig:lemniscate}(e).
These results are surprisingly accurate, given the absence of aerodynamic priors and the maximum tilt angle exceeding 80$^\circ$.
\subsubsection{Fast Transitions}
\Para{experiments-11}
To assess the framework under aggressive operating conditions, we conducted rapid transition experiments in comparison to a conventional linear transition baseline~\cite{lyu-cas}.
To ensure a fair comparison, both methods utilize the same low-level controller. The level-flight speed is set to 12\,m/s with a nominal 15$^\circ$ pitch angle, requiring a minimum of 75$^\circ$ pitch motion during transition. Note that the baseline~\cite{lyu-cas} utilizes aerodynamic priors and regulates altitude, whereas the proposed method achieves 3D position control without aerodynamic priors.
Fig.~\ref{fig:rapid-transition} illustrates the forward and backward transition results.
By tuning the tangential acceleration constraint~\eqref{eq:feas-contr-at} progressively, our method completes the forward transition in just 2.0\,s. Furthermore, it executes a sharp pitch motion for rapid deceleration, completing the backward transition in only 1.8\,s. Most notably, even during this most aggressive transition, our prior-free method achieves a significantly lower height MaxAE. Specifically, our method reduces height errors to 0.54\,m (forward) and 0.25\,m (backward), consistently outperforming the baseline (1.15\,m and 1.00\,m) in both aggressiveness and precision.
\section{Conclusion and Future Work} \label{sec:conclusion}
\Para{conclusion-01}
 In this work, we present a trajectory planning and tracking framework for coordinated tail-sitter flights that does not require airframe-specific aerodynamic priors. Simulations confirm that the proposed method closely tracks the high-fidelity MPC upper bound over the tested flight regimes, while real-world field experiments further validate the framework's efficacy. At extreme speeds, however, high-fidelity aerodynamic modeling remains important for reducing worst-case tracking errors. Since the approach does not require airframe-specific aerodynamic priors, it can be readily adapted to various configurations and easily deployed. Additionally, it can serve as a reliable initial controller for collecting high-quality flight data, facilitating the subsequent design of high-fidelity model-based planning and tracking.
\Para{conclusion-02}
Wind-disturbance simulations further characterize the framework's empirical operating range. Over the tested conditions, the framework maintains $\mathrm{MaxAE}\leq1$\,m under moderate tailwinds but degrades under even mild headwind and crosswind disturbances. The worst degradation occurs under combined headwind and crosswind, where the vehicle enters the low-AoA high-stiffness regime while sideslip simultaneously violates the coordinated-flight constraint. Future work will extend the framework to be wind-aware while retaining its aerodynamic prior-free design.
\bibliography{references}

\begin{thebibliography}{10}
\providecommand{\url}[1]{#1}
\csname url@samestyle\endcsname
\providecommand{\newblock}{\relax}
\providecommand{\bibinfo}[2]{#2}
\providecommand{\BIBentrySTDinterwordspacing}{\spaceskip=0pt\relax}
\providecommand{\BIBentryALTinterwordstretchfactor}{4}
\providecommand{\BIBentryALTinterwordspacing}{\spaceskip=\fontdimen2\font plus
\BIBentryALTinterwordstretchfactor\fontdimen3\font minus
  \fontdimen4\font\relax}
\providecommand{\BIBforeignlanguage}[2]{{%
\expandafter\ifx\csname l@#1\endcsname\relax
\typeout{** WARNING: IEEEtran.bst: No hyphenation pattern has been}%
\typeout{** loaded for the language `#1'. Using the pattern for}%
\typeout{** the default language instead.}%
\else
\language=\csname l@#1\endcsname
\fi
#2}}
\providecommand{\BIBdecl}{\relax}
\BIBdecl

\bibitem{lu2024trajectory}
G.~Lu, Y.~Cai, N.~Chen, F.~Kong, Y.~Ren, and F.~Zhang, ``Trajectory generation
  and tracking control for aggressive tail-sitter flights,'' \emph{Int. J.
  Robot. Res.}, vol.~43, no.~3, pp. 241--280, 2024.

\bibitem{lu2025autonomous}
G.~Lu, Y.~Ren, F.~Zhu, H.~Li, R.~Xue, Y.~Cai, X.~Lyu, and F.~Zhang,
  ``Autonomous tail-sitter flights in unknown environments,'' \emph{IEEE Trans.
  Robot.}, vol.~41, pp. 1098--1117, 2025.

\bibitem{eth-full}
S.~Verling, B.~Weibel, M.~Boosfeld, K.~Alexis, M.~Burri, and R.~Siegwart,
  ``Full attitude control of a {VTOL} tailsitter {UAV},'' in \emph{Proc. IEEE
  Int. Conf. Robot. Autom.}, 2016, pp. 3006--3012.

\bibitem{lyu-cas}
X.~Lyu, H.~Gu, J.~Zhou, Z.~Li, S.~Shen, and F.~Zhang, ``Simulation and flight
  experiments of a quadrotor tail-sitter vertical take-off and landing unmanned
  aerial vehicle with wide flight envelope,'' \emph{Int. J. Micro Air
  Vehicles}, vol.~10, no.~4, pp. 303--317, 2018.

\bibitem{zhou-scp-att}
J.~Zhou, X.~Lyu, Z.~Li, S.~Shen, and F.~Zhang, ``A unified control method for
  quadrotor tail-sitter {UAV}s in all flight modes: Hover, transition, and
  level flight,'' in \emph{Proc. IEEE/RSJ Int. Conf. Intell. Robots Syst.},
  2017, pp. 4835--4841.

\bibitem{eth-flywing}
R.~Ritz and R.~D'Andrea, ``A global controller for flying wing tailsitter
  vehicles,'' in \emph{Proc. IEEE Int. Conf. Robot. Autom.}, 2017, pp.
  2731--2738.

\bibitem{quanquan-liftwing}
Q.~Quan, S.~Wang, and W.~Gao, ``Lifting-wing quadcopter modeling and unified
  control,'' \emph{J. Guid. Control Dyn.}, vol.~48, no.~3, pp. 689--699, 2025.

\bibitem{indi-smeur}
E.~J.~J. Smeur, M.~Bronz, and G.~C. H.~E. de~Croon, ``Incremental control and
  guidance of hybrid aircraft applied to a tailsitter unmanned air vehicle,''
  \emph{J. Guid. Control Dyn.}, vol.~43, no.~2, pp. 274--287, 2020.

\bibitem{mit_estimation}
E.~Tal and S.~Karaman, ``Global incremental flight control for agile
  maneuvering of a tailsitter flying wing,'' \emph{J. Guid. Control Dyn.},
  vol.~45, no.~12, pp. 2332--2349, 2022.

\bibitem{rohr2024unified}
D.~Rohr, O.~Andersson, N.~Lawrance, T.~Stastny, and R.~Siegwart, ``Unified
  guidance and jerk-level dynamic inversion for accurate position control of
  hybrid uavs,'' \emph{IEEE Trans. Robot.}, vol.~41, pp. 708--728, 2025.

\bibitem{xuwei-acc}
W.~Xu and F.~Zhang, ``Learning pugachev's cobra maneuver for tail-sitter {UAV}s
  using acceleration model,'' \emph{IEEE Robot. Autom. Lett.}, vol.~5, no.~2,
  pp. 3452--3459, 2020.

\bibitem{RAL-ducted-fan}
Z.-H. Cheng and H.-L. Pei, ``Transition analysis and practical flight control
  for ducted fan fixed-wing aerial robot: Level path flight mode transition,''
  \emph{IEEE Robot. Autom. Lett.}, vol.~7, no.~2, pp. 3106--3113, 2022.

\bibitem{song2023reaching}
Y.~Song, A.~Romero, M.~Müller, V.~Koltun, and D.~Scaramuzza, ``Reaching the
  limit in autonomous racing: Optimal control versus reinforcement learning,''
  \emph{Science Robotics}, vol.~8, no.~82, p. eadg1462, 2023.

\bibitem{kaufmann2023champion}
E.~Kaufmann, L.~Bauersfeld, A.~Loquercio, M.~M{\"u}ller, V.~Koltun, and
  D.~Scaramuzza, ``Champion-level drone racing using deep reinforcement
  learning,'' \emph{Nature}, vol. 620, no. 7976, pp. 982--987, 2023.

\bibitem{skydreamer2025}
A.~Verraest, S.~Bahnam, R.~Ferede, G.~de~Croon, and C.~De~Wagter, ``Skydreamer:
  Interpretable end-to-end vision-based drone racing with model-based
  reinforcement learning,'' \emph{arXiv preprint arXiv:2510.14783}, 2025.

\bibitem{differentiable-physics}
Y.~Zhang, Y.~Hu, Y.~Song, D.~Zou, and W.~Lin, ``Learning vision-based agile
  flight via differentiable physics,'' \emph{Nature Machine Intelligence},
  vol.~7, no.~6, pp. 954--966, 2025.

\bibitem{pla-dic-collocation}
A.~J. Barry, T.~Jenks, A.~Majumdar, H.-T. Lin, I.~G. Ros, A.~A. Biewener, and
  R.~Tedrake, ``Flying between obstacles with an autonomous knife-edge
  maneuver,'' in \emph{Proc. IEEE Int. Conf. Robot. Autom.}, 2014, pp.
  2559--2565.

\bibitem{mp_library_2019}
J.~M. Levin, M.~Nahon, and A.~A. Paranjape, ``Real-time motion planning with a
  fixed-wing {UAV} using an agile maneuver space,'' \emph{Auton. Robots},
  vol.~43, no.~8, pp. 2111--2130, 2019.

\bibitem{mp_library_2020}
D.~H. Lee, C.-J. Kim, M.~J. Heo, J.~W. Hwang, H.~G. Lyu, and J.~Y. Lee,
  ``Development of real-time maneuver library generation technique for
  implementing tactical maneuvers of fixed-wing aircraft,'' \emph{Int. J.
  Aerosp. Eng.}, vol. 2020, pp. 1--12, 2020.

\bibitem{turkey2023}
S.~Aslan, S.~Demirci, T.~Oktay, and E.~Yesilbas, ``Percentile-based adaptive
  immune plasma algorithm and its application to engineering optimization,''
  \emph{Biomimetics}, vol.~8, no.~6, p. 486, 2023.

\bibitem{turkey2023uav}
S.~Aslan and T.~Oktay, ``Path planning of an unmanned combat aerial vehicle
  with an extended-treatment-approach-based immune plasma algorithm,''
  \emph{Aerospace}, vol.~10, no.~5, p. 487, 2023.

\bibitem{minco2022}
Z.~Wang, X.~Zhou, C.~Xu, and F.~Gao, ``Geometrically constrained trajectory
  optimization for multicopters,'' \emph{IEEE Trans. Robot.}, vol.~38, no.~5,
  pp. 3259--3278, 2022.

\bibitem{yu2025top}
J.~Yu, N.~Chen, G.~Liu, C.~Xu, F.~Gao, and Y.~Cao, ``Top: Trajectory
  optimization via parallel optimization towards constant time complexity,''
  \emph{IEEE Robot. Autom. Lett.}, vol.~10, no.~12, pp. 13\,153--13\,160, 2025.

\bibitem{df_co}
J.~Hauser and R.~Hindman, ``Aggressive flight maneuvers,'' in \emph{Proc. IEEE
  Control Decis. Conf.}, 1997, pp. 4186--4191.

\bibitem{df-low}
T.~Liu, M.~Wang, Y.~Niu, J.~Li, and H.~Zhou, ``Low-cost differential flatness
  identification for trajectory planning and tracking of small fixed-wing
  {UAV}s in dense environments,'' in \emph{Proc. Int. Conf. Unmanned Aircr.
  Syst.}, Jun 2024, pp. 905--910.

\bibitem{df-longitudinal}
K.~McIntosh, J.~Reddinger, S.~Mishra, and D.~Zhao, ``Optimal trajectory
  generation for transitioning quadrotor biplane tailsitter using differential
  flatness,'' in \emph{Proc. Vert. Flight Soc. 77th Annu. Forum}, 2021, pp.
  1--9.

\bibitem{df-lift-wing}
S.~Wang, W.~Gao, and Q.~Quan, ``Differential flatness of lifting-wing
  quadcopters subject to drag and lift for accurate tracking,'' \emph{IEEE
  Trans. Ind. Electron.}, vol.~71, no.~10, pp. 12\,664--12\,673, 2024.

\bibitem{mit_traj2022}
E.~Tal, G.~Ryou, and S.~Karaman, ``Aerobatic trajectory generation for a {VTOL}
  fixed-wing aircraft using differential flatness,'' \emph{IEEE Trans. Robot.},
  vol.~39, no.~6, pp. 4805--4819, 2023.

\bibitem{phi2019}
L.~R. Lustosa, F.~Defaÿ, and J.-M. Moschetta, ``Global singularity-free
  aerodynamic model for algorithmic flight control of tail sitters,'' \emph{J.
  Guid. Control Dyn.}, vol.~42, no.~2, pp. 303--316, 2019.

\bibitem{lewis-line}
A.~S. Lewis and M.~L. Overton, ``Nonsmooth optimization via quasi-{N}ewton
  methods,'' \emph{Mathematical Programming}, vol. 141, no. 1-2, pp. 135--163,
  2013.

\bibitem{lbfgs}
D.~C. Liu and J.~Nocedal, ``On the limited memory {BFGS} method for large scale
  optimization,'' \emph{Mathematical Programming}, vol.~45, no. 1-3, pp.
  503--528, 1989.

\bibitem{autodiff}
\BIBentryALTinterwordspacing
A.~M.~M. Leal, ``Autodiff, a modern, fast and expressive {C++} library for
  automatic differentiation,'' \texttt{https://autodiff.github.io}, 2018.
  [Online]. Available: \url{https://autodiff.github.io}
\BIBentrySTDinterwordspacing

\bibitem{hpipm}
G.~Frison and M.~Diehl, ``{HPIPM}: a high-performance quadratic programming
  framework for model predictive control,'' \emph{IFAC-PapersOnLine}, vol.~53,
  no.~2, pp. 6563--6569, 2020.

\bibitem{acados}
R.~Verschueren, G.~Frison, D.~Kouzoupis, J.~Frey, N.~v. Duijkeren, A.~Zanelli,
  B.~Novoselnik, T.~Albin, R.~Quirynen, and M.~Diehl, ``Acados—a modular
  open-source framework for fast embedded optimal control,'' \emph{Mathematical
  Programming Computation}, vol.~14, no.~1, pp. 147--183, 2022.

\bibitem{casadi}
J.~A.~E. Andersson, J.~Gillis, G.~Horn, J.~B. Rawlings, and M.~Diehl,
  ``{CasADi}: A software framework for nonlinear optimization and optimal
  control,'' \emph{Mathematical Programming Computation}, vol.~11, no.~1, pp.
  1--36, 2019.

\bibitem{MaAero}
Z.~Ma, E.~J.~J. Smeur, and G.~C. H.~E. de~Croon, ``Wind tunnel tests of a wing
  at all angles of attack,'' \emph{Int. J. Micro Air Veh.}, vol.~14, 2022, art.
  no. 17568293221110931.

\end{thebibliography}
\begin{IEEEbiography}[{\includegraphics[width=1in,height=1.25in,clip,keepaspectratio]{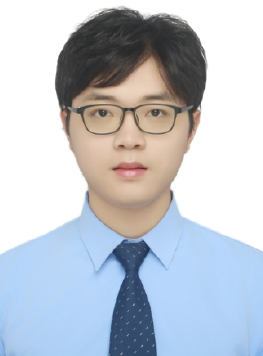}}]{Erchao Rong}
received the B.Eng. degree in Intelligent Science and Technology (2022) and the M.Eng. degree in Control Science and Engineering (2025) from Sun Yat-sen University, Guangzhou, China. During his master's studies, he focused on trajectory planning and control of VTOL UAVs. He is currently working on bringing agentic autonomous flight to aerial vehicles.
\end{IEEEbiography}
\begin{IEEEbiography}[{\includegraphics[width=1in,height=1.25in,clip,keepaspectratio]{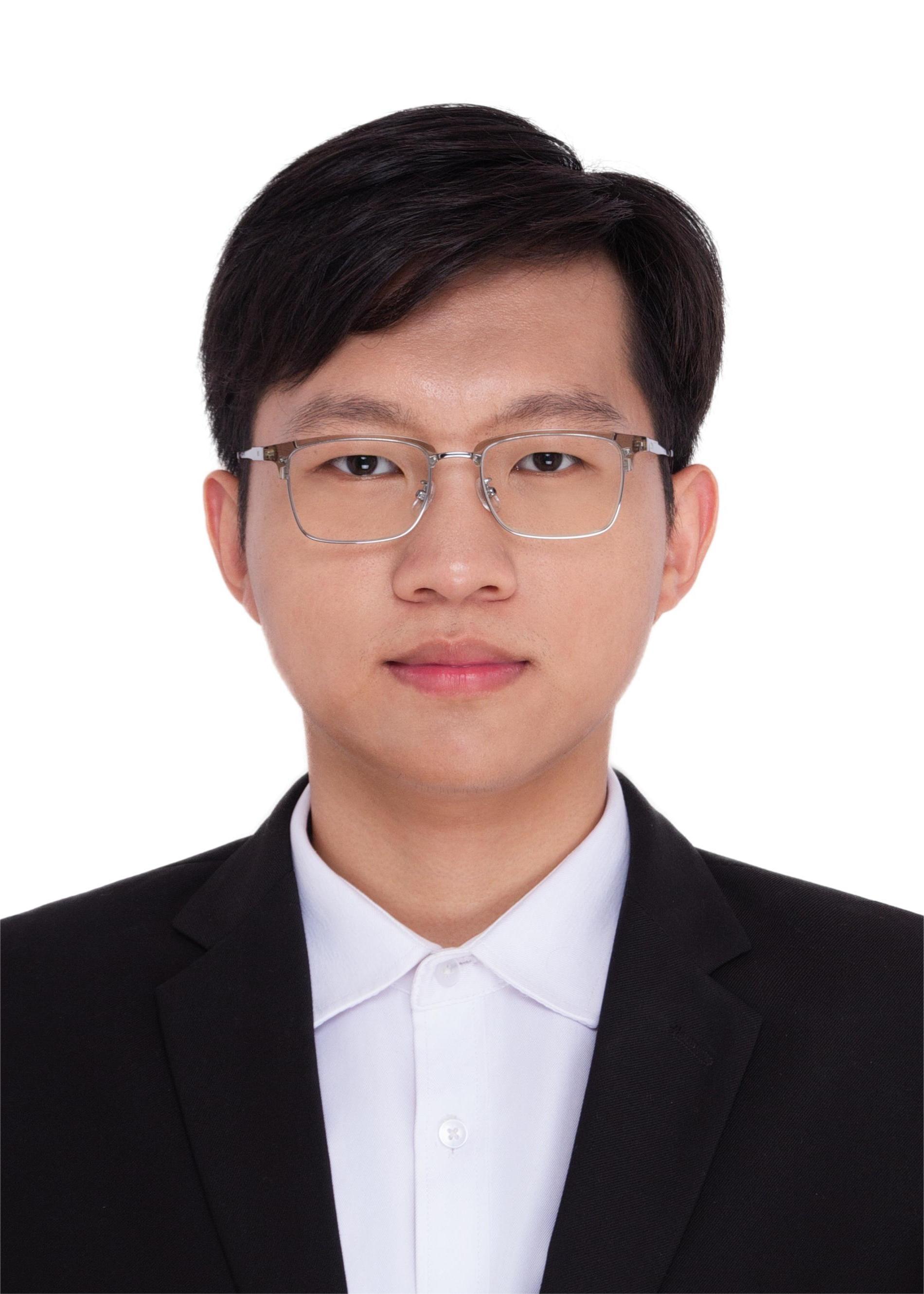}}]{Zihao Liu}
received the B.Eng. degree in intelligent science and technology from Sun Yat-sen University, Guangzhou, China, in 2025. He is currently pursuing the M.S. degree in control science and engineering at the same university. His research interests include planning and control of VTOL UAVs and learning-based UAV flight control.
\end{IEEEbiography}
\begin{IEEEbiography}[{\includegraphics[width=1in,height=1.25in,clip,keepaspectratio]{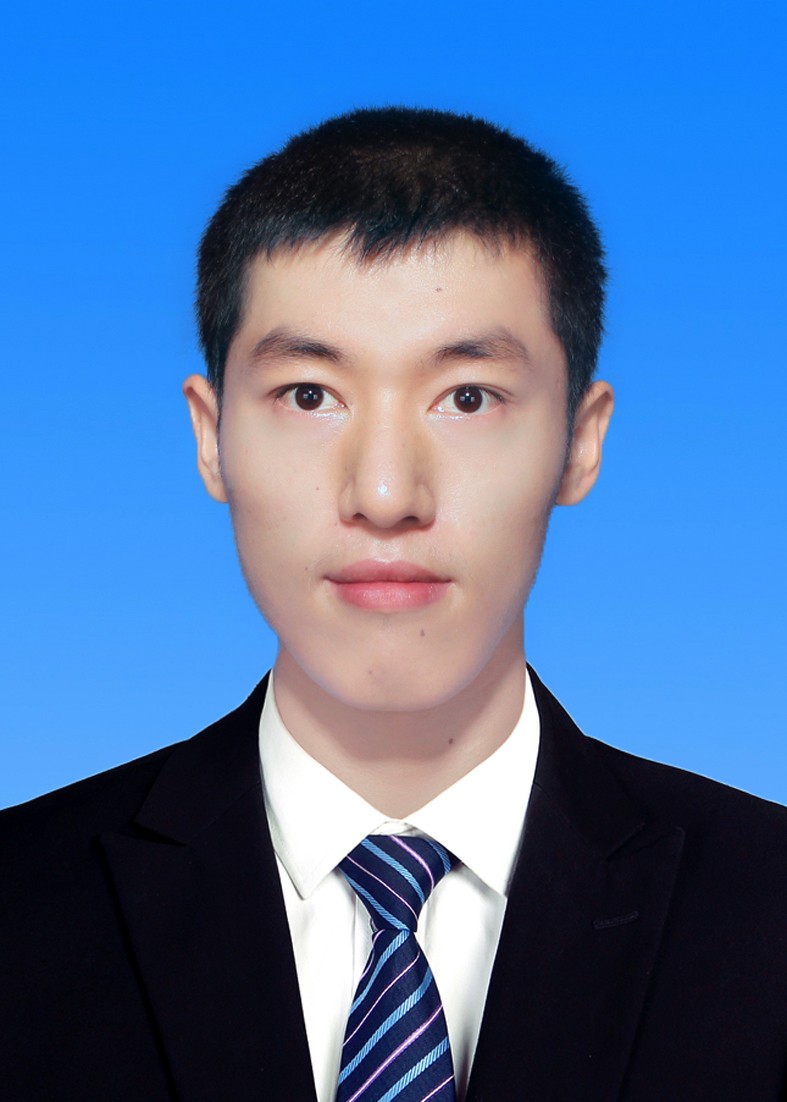}}]{Junning Liang}
received the M.Eng. degree in Mechanical Engineering from Sun Yat-sen University, Guangzhou, China, in 2024.
He is currently pursuing the Ph.D. degree at Sun Yat-sen University, Guangzhou, China.
His research interests include flight control of aerial robots.
\end{IEEEbiography}
\begin{IEEEbiography}[{\includegraphics[width=1in,height=1.25in,clip,keepaspectratio]{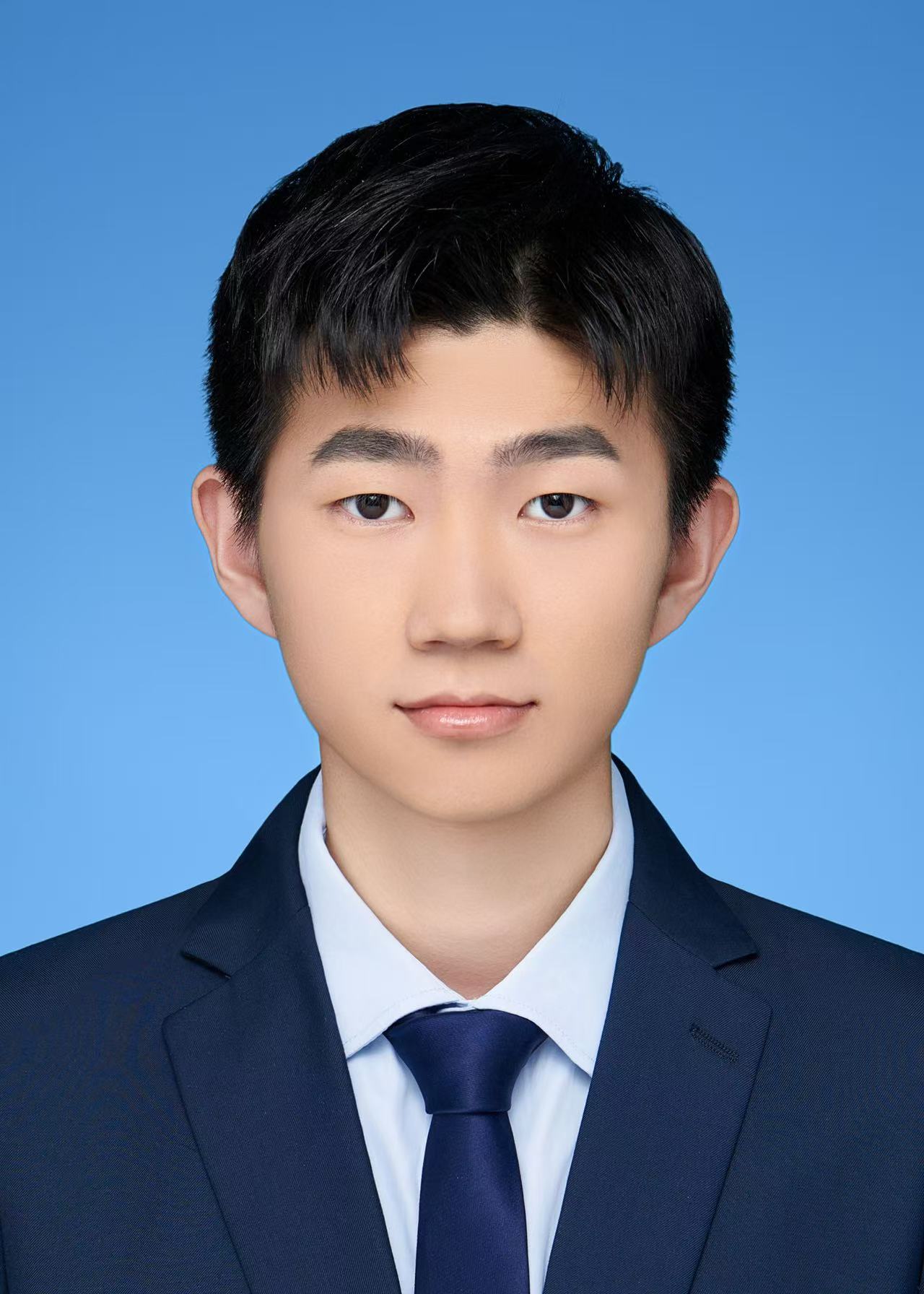}}]{Jianguo Wang}
received the B.Eng. degree in automation from the School of Control
Science and Engineering, Shandong University, Jinan, China, in 2024. He is currently pursuing an M.Eng. degree with the School of Intelligent Systems Engineering, Sun Yat-sen University, Guangzhou, China. His research interests include the control and trajectory planning of robotic systems.
\end{IEEEbiography}
\begin{IEEEbiography}[{\includegraphics[width=1in,height=1.25in,clip,keepaspectratio]{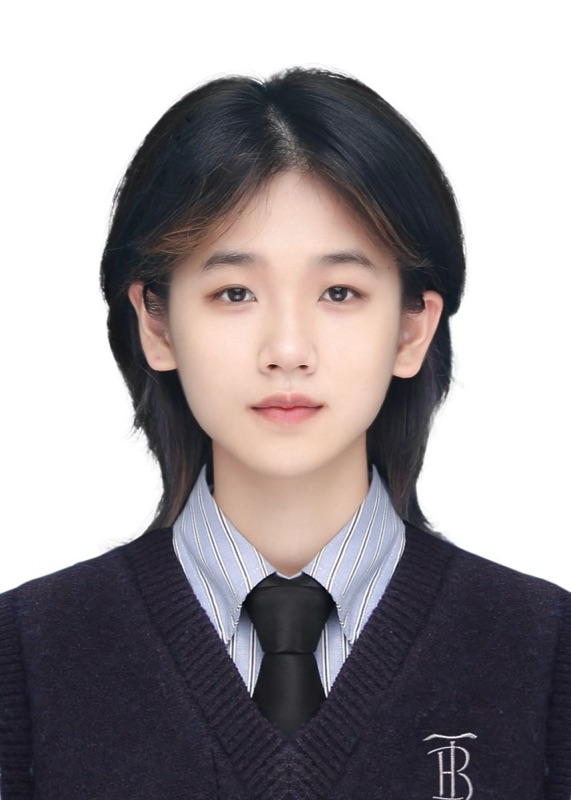}}]{Xiao Jie}
obtained the B.Eng. degree from the School of Intelligent Systems Engineering, Sun Yat-sen University, Guangzhou, China in 2025. She is currently pursuing an M.Sc. degree at the Faculty of Engineering at the University of Hong Kong, with research interests in the motion control of VTOL UAVs and air-ground coordination.
\end{IEEEbiography}
\begin{IEEEbiography}[{\includegraphics[width=1in,height=1.25in,clip,keepaspectratio]{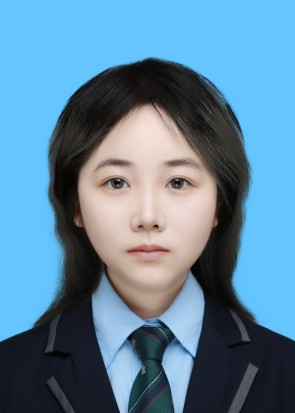}}]{Haoran Fu}
received the B.S. degree in Robotics Engineering from Northeastern University, Shenyang, China, in 2022. She is currently a master's student at the School of Intelligent Systems Engineering, Sun Yat-sen University, Guangzhou, China. Her research interests focus on UAV planning and control.
\end{IEEEbiography}
\begin{IEEEbiography}[{\includegraphics[width=1in,height=1.25in,clip,keepaspectratio]{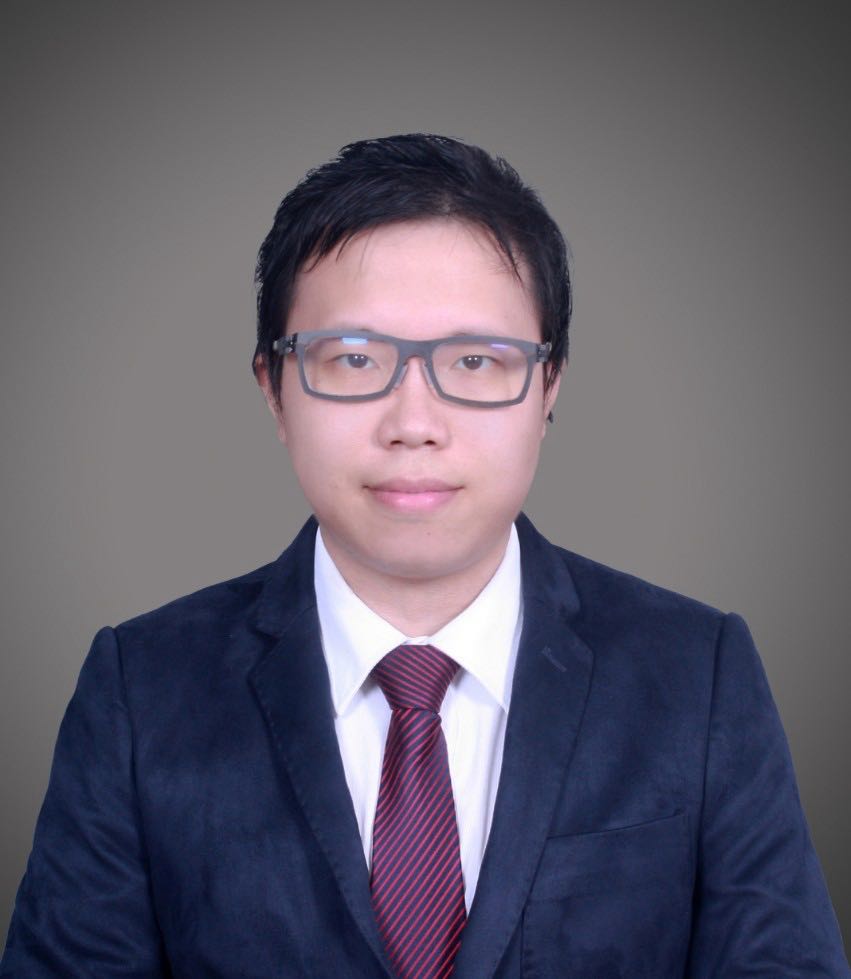}}]{Ziliang Chen}
 is an Assistant Researcher at Pengcheng Laboratory, Shenzhen, China. He received the Ph.D. degree in Computer Science and Technology from Sun Yat-sen University, Guangzhou, China, in 2021, and the B.S. degree in Mathematics from Sun Yat-sen University in 2010.
\end{IEEEbiography}
\begin{IEEEbiography}[{\includegraphics[width=1in,height=1.25in,clip,keepaspectratio]{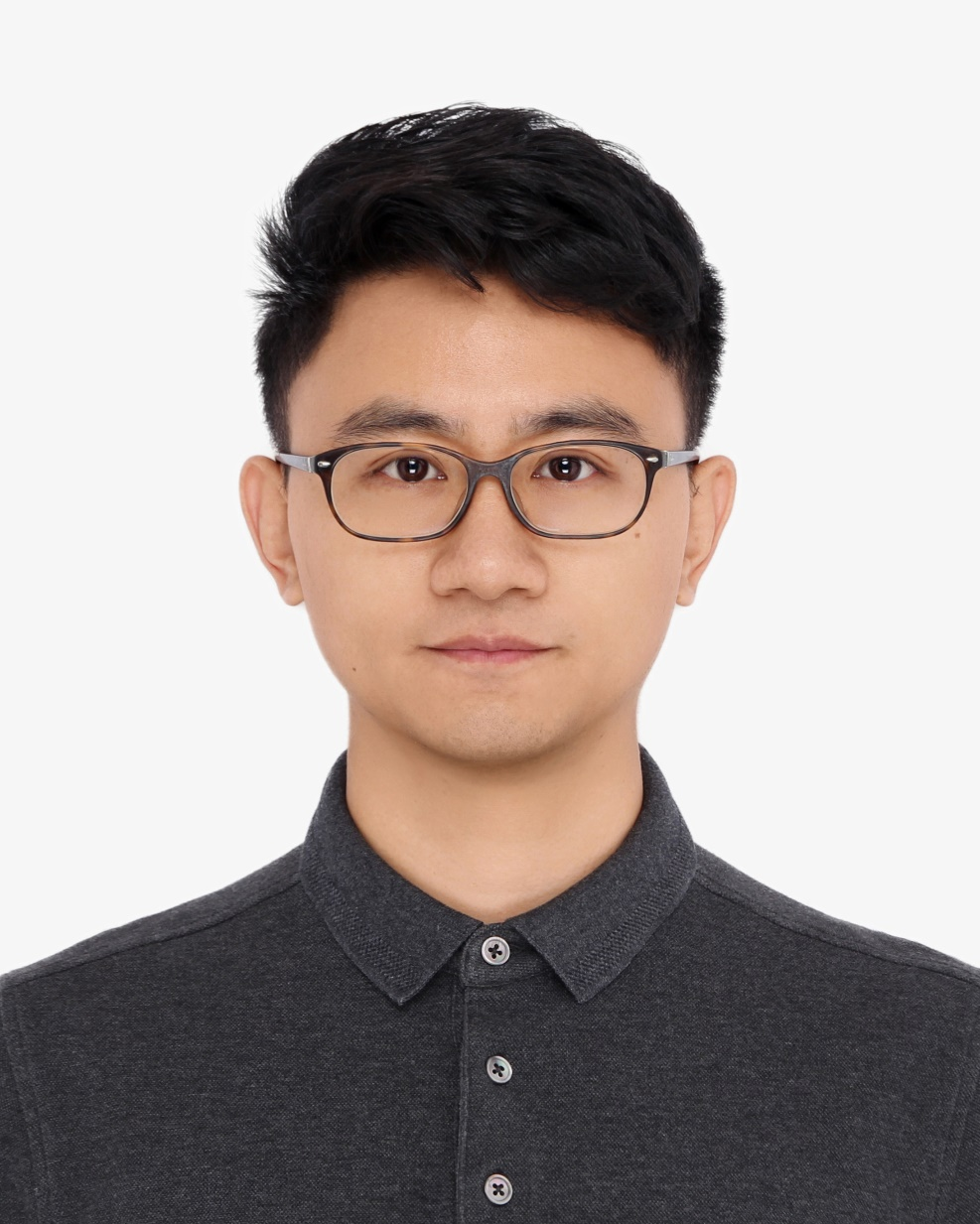}}]{Ximin Lyu}
received his B.Eng. and M.Phil. degrees in Aircraft Manufacturing from Harbin Institute of Technology, Harbin, China, in 2012 and 2014, respectively. He received his Ph.D. in Electronic and Computer Engineering from the Hong Kong University of Science and Technology, Hong Kong, China, in 2019.
In 2018, he served as a Senior Flight Control Researcher at Da Jiang Innovation (DJI) Technology Company in China. Since 2021, he has been an Associate Professor with the Department of Intelligent Systems Engineering, Sun Yat-sen University, Guangzhou, China.
His research interests encompass robotics and controls, with a specific focus on UAV/UGV design, control, and planning.
\end{IEEEbiography}
\end{document}